\pdfoutput=1
\documentclass[11pt]{article}

\usepackage[final]{acl}

\usepackage{times}
\usepackage{latexsym}

\usepackage[T1]{fontenc}
\usepackage[utf8]{inputenc}

\usepackage{microtype}

\usepackage{inconsolata}

\usepackage{graphicx}
\usepackage{booktabs}
\usepackage{arydshln}
\usepackage{multirow}
\usepackage{makecell}
\usepackage{array}
\usepackage{amsmath}
\usepackage{kotex}
\usepackage{tcolorbox}
\usepackage{amsfonts} 
\usepackage{subcaption}

\title{RV-HATE: Reinforced Multi-Module Voting for Implicit Hate Speech Detection}

\author{
 \textbf{Yejin Lee},
 \textbf{Hyeseon An},
 \textbf{Yo-Sub Han}\thanks{Corresponding author.}
 \\
  Yonsei University, Seoul, Republic of Korea,
\\
   \texttt{\{%
   \href{mailto:ssgyejin@yonsei.ac.kr}{ssgyejin},%
   \href{mailto:hsan@yonsei.ac.kr}{hsan},%
   \href{mailto:emmous@yonsei.ac.kr}{emmous}%
   \}@yonsei.ac.kr}
}

\begin{document}
\maketitle
\begin{abstract}
Hate speech remains prevalent in human society and 
continues to evolve in its forms and expressions.
Modern advancements in the internet and online anonymity 
accelerate its rapid spread and complicate its detection.
However, hate speech datasets exhibit diverse characteristics
primarily because they are constructed from different sources
and platforms, each reflecting different linguistic styles and social contexts.
Despite this diversity, prior studies on hate speech detection
often rely on fixed methodologies without adapting to data-specific features.
We introduce RV-HATE, a detection framework designed to
account for the dataset-specific characteristics of each hate speech dataset.
RV-HATE consists of multiple specialized modules,
where each module focuses on distinct linguistic or contextual features
of hate speech.
The framework employs reinforcement learning to optimize weights that determine the contribution of each module for a given dataset.
A voting mechanism then aggregates the module outputs to produce
the final decision.
RV-HATE offers two primary advantages: (1)~it improves
detection accuracy by tailoring the detection process  to dataset-specific attributes, and
(2)~it also provides interpretable insights into the distinctive features of each dataset.
Consequently, our approach effectively addresses implicit hate speech and
achieves superior performance compared to conventional static methods. Our code is available at \url{https://github.com/leeyejin1231/RV-HATE}.

\end{abstract}

\section{Introduction}

\textbf{Warning}: \textit{this paper contains content that may be offensive and upsetting.}

% Problem: (Implicit) hate speech detection은 성능 올리기 어렵다.
% Motivation: 기존 연구들은 (xxx,yyy) 데이터셋 별 특징을 고려하지 않았다. 특히 hate speech dataset은 주관적인 데이터로 유명함에도 불구하고!!! 기존에는 데이터셋별 차이점을 죽였다면.
% Approach: 기존의 이런 특성을 무시하는 방법들과 달리 우리는 이 다른 데이터셋별 특징점을 오히려 살려보겠다!

%% 현재
%% 문제
%% 기존연구와 문제
%% 그래서 우리는 ~ 했다
%% 우리거 설명
%% 그래서 우리가 발견한거 분석 설명
%% 결론 (수치 제시할 필요는 없음)

% 온라인 사용의 증가로 offensive speech 등장
Online platforms continue to grow rapidly and this growth increases the prevalence of hate speech~\citep{madriaza2025exposure}.
% hate speech의 정의
Hate speech refers to language that promotes hatred, discrimination,
or violence toward a specific group or community---gender,
race, religion, nationality, or other identities~\citep{poletto2021}.
It is typically categorized into two types: explicit and implicit.
While explicit hate speech can be easily identified via explicit abusive 
expressions or lexicons~\citep{WaseemDWW17,CaselliBMKG20,OcampoSCV23},
implicit hate speech remains challenging due to its subtle and
context-dependent nature.
% While explicit hate speech expresses the harmful expressions or slurs in a direct,
% implicit hate speech relies on contextual cues in a more subtle or indirect way.
For these reasons, detecting hate speech has become a critical task.
% however it remains challenging due to the complex and context-dependent nature
% 각 데이터별 예문과 특징
Additionally, hate speech datasets exhibit diverse linguistic 
and contextual variation, 
primarily due to their construction from diverse sources and platforms 
that reflect different language conventions and social dynamics.
As a result, these datasets vary in linguistic style,
degree of implicitness, and annotation criteria.
% IHC~\citep{ElSheriefZMASCY21} contains mostly implicit hate speech, where the hateful intent appears indirectly through subtle expressions or
% cultural references.
% IHC comes from Twitter and often includes grammatically incorrect
% or fragmented texts, which produce many broken sentences.
% (e.g., "only solution = white revolution") 
% SBIC~\citep{SapGQJSC20} focuses on offensive speech.
% Offensive speech covers a broader concept than hate speech
% and includes expressions that insult, offend or cause discomfort
% without necessarily targeting a specific group~\cite{poletto2021}.
% (e.g., "Molly make bitches crazy")
% DYNA~\citep{VidgenTWK20} combines human and machine-generated content and
% provides relatively more stable and less noisy text
% compared to other datasets.
% (e.g., "You only played yourself there, fucking retard")
% Hateval further suffers from ambiguous labeling between
% offensive and hateful content. 
% (e.g., "that bitch wanna act funny bitch ima act hysterical")
% A more detailed analysis appears in Section~\ref{sec:approach}.
% 데이터 마다 특징이 다름
% These datasets show distinct differences in linguistic style,
% implicitness and labeling criteria.
Therefore, robust hate speech detection methods must account for 
the dataset-specific characteristics of individual datasets.

% 기존의 연구는 ~~, 이러한 데이터셋의 특징들을 무시하고 하나로 시행
Recent research has focused on developing methods for hate speech detection.
For instance, \citet{AhnKKH24} employed clustering in the sentence embedding space to identify representative samples for contrastive learning,
and \citet{KimJPPH24} employed an additional queue for hard negative samples beyond the batch-level.
Although prior studies have made progress in implicit hate speech detection, 
they often overlook dataset-specific features and distinctive labeling criteria.
We must design methods that account for the distinct characteristics of each dataset to effectively address these limitations.
% 그래서 VR-HATE 도입

We propose a reinforced multi-module voting method for implicit hate speech detection (RV-HATE).
RV-HATE consists of multiple modules
designed to capture the unique properties of each dataset 
and employs a reinforced voting mechanism that adaptively optimizes their contributions based on dataset-specific characteristics.
% VR-HATE 설명
RV-HATE consists of four modules ($\mathtt{M_0}$ -- $\mathtt{M_3}$): The
$\mathtt{M_0}$~module serves as the base module, capturing the context of hate speech
by using the cosine similarity
during clustering-based contrastive learning. 
The $\mathtt{M_1}$~module tags the hate targets, thereby enabling more precise discrimination of hate speech. 
The $\mathtt{M_2}$~module removes the outliers in clusters of the dataset  
to guarantee the quality of data
and the $\mathtt{M_3}$~module utilizes hard negative samples during the contrastive learning
to provide a clear decision boundary.
Each module is constructed by augmenting the base module $\mathtt{M_0}$
with its corresponding functionality.
We fine-tune the four classifier models with each module,
and employ these classifiers in a voting process,
where reinforcement learning dynamically assigns dataset-specific weights
to each classifier.
% This approach enables RV-HATE to capture the linguistic and contextual nuances of each dataset.

% 데이터셋 분석 (알게된거) 설명
% 결론 (우리 짱이다)
Through this voting mechanism,
RV-HATE not only outperforms previous state-of-the-art (SOTA) models by an average of 1.8\%p,
but also provides interpretability by revealing how each module contributes under varying dataset characteristics.
While an improvement of 1--2\% may appear incremental, 
such gains are particularly significant in hate speech detection,
where performance typically plateaus around the 80\% range~\cite{AhnKKH24, KimJPPH24}.
Our approach highlights the importance of dataset-specific strategy
in hate speech detection and demonstrates that employing multi-modules can
both improve detection performance and provide explainable insights into the characteristics of implicit hate speech datasets.

Our main contributions are as follows.
\begin{itemize}
    \item We develop four specialized modules designed to capture diverse dataset-specific characteristics of implicit hate speech.
    \item We present a reinforcement learning-based voting mechanism that assigns dataset-specific weights to each module, enabling adaptive combination of module predictions.
    \item  We improve the detection performance across multiple benchmarks and provide interpretable insights into how different modules contribute under the dataset characteristics.
\end{itemize}

\begin{figure*}[ht!]
    \centering
    \includegraphics[width=\textwidth]{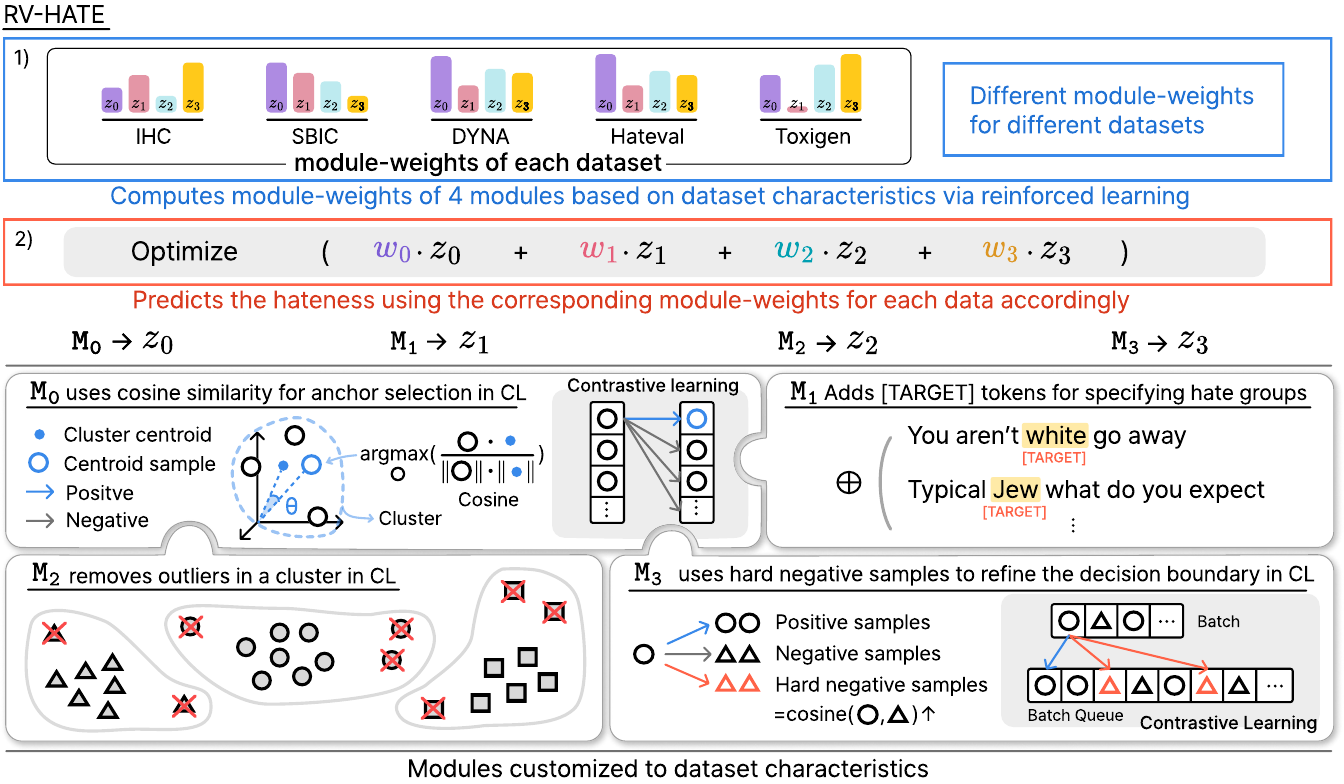}
         \caption{Overall workflow of RV-HATE. The method processes implicit hate speech data through four modules
         $\mathtt{M_0}$ (Sec.~\ref{subsec:module1}),
         $\mathtt{M_1}$ (Sec.~\ref{subsec:module2}),
         $\mathtt{M_2}$ (Sec.~\ref{subsec:module3}), and
         $\mathtt{M_3}$ (Sec.~\ref{subsec:module4}).
         Reinforcement learning is employed to determine the optimal
         weights for these modules in the voting process.}
    \label{fig:overview}
\end{figure*}

\section{Related Work}

\subsection{Implicit Hate Speech Detection}
Hate speech detection plays a crucial role in mitigating online toxicity
and preventing the spread of harmful communication~\citep{gandhi2024hate, LEE201822}.
Implicit hate speech expresses hateful or discriminatory intent indirectly, 
often relying on context instead of explicit slurs or offensive sentences~\citep{WeberVZS20}.
This subtle aspect makes implicit hate speech more challenging to detect than explicit hate speech. 
There are a few datasets for implicit hate speech detection.
\citet{ElSheriefZMASCY21} introduced the Implicit Hate Corpus~(IHC)
that captures subtle and indirect hate speech relying on contextual cues and implicit stereotypes. 
The dataset contains implications for each hateful sentence, which provide explanations of their implied meanings.
\citet{HartvigsenGPSRK22} focused on adversarial non-toxic counterfactuals to evaluate model generalization in hate speech detection 
and incorporated LLM-generated neutral sentences.
\citet{LeeJMJCKO24} constructed a cross-cultural English hate speech dataset
and analyzed how cultural background influences hate speech annotations across different countries.
In addition to dataset development,
researchers tackled the
implicit hate speech problem
by investigating its linguistic structures 
and the difficulties posed by context-dependent or culturally grounded
expressions~\citep{FortunaCW21,DavaniAKD23,OcampoSCV23}.
% There are different training methodologies for improving the performance of detecting implicit hate speech.

\citet{HuangKA23} assessed the capability of LLMs to detect implicit hate speech by generating 
concise natural language explanations.
% \citet{GhoshSCTKM23} employed a hyperbolic network 
% to effectively encode diverse external contexts, 
% thereby enhancing implicit hate speech detection in online discussions.
\citet{ParkKJPH24} introduced a multi-agent-based debate simulation
framework that generates diverse perspectives on implicit hate speech.
These approaches demonstrate the significance of improving the robustness of implicit hate speech detection models.
\citet{KimAKH25} presented a CONELA, a data refinement strategy that utilizes human agreement and training dynamics to improve generalization in implicit hate speech detection.
\citet{LeeAH25} proposed a target-aware attention framework that models interactions between explicit and implicit targets and their context to improve the ability to detect implicit hate speech.

\subsection{Hate Speech Detection with Contrastive Learning}
Researchers notice that 
contrastive learning is effective for detecting implicit hate speech by 
distinguishing the subtle semantic nuances and context-dependent cues that
characterize such language.
% detection to improve the performance 
% of models to detect hate speech from neutral or subtly harmful expressions. 
\citet{KimPH22} applied contrastive learning using implication data 
to detect implicit hate speech from neutral text. 
\citet{KimPNH23} proposed an augmentation approach that 
utilizes machine-generated data to enhance implicit hate speech detection.
\citet{HuangU24} employed a revisiting supervised contrastive learning method for subtle hate speech detection.
\citet{KimJPPH24} proposed a negative sampling strategy
using momentum contrastive learning with an additional queue. 
\citet{Jiang25} proposed an approach that applies the causal inference method
to refine contrastive learning.

\citet{AhnKKH24} proposed SharedCon, a SOTA model
for implicit hate speech detection.
The model employs clustering-based contrastive learning to improve contextual
representation and model robustness.
In contrastive learning, the model learns to bring similar
samples closer in the representation space using an anchor.
SharedCon selects the sample closest to each cluster center
as an anchor, allowing the model to effectively learn shared semantic
patterns across data.
%Motivated by them, we also employ a clustering algorithm in our framework to guarantee
%the use of shared information among data points.

% \subsection{Ensemble with Reinforcement Learning}

% Traditional ensemble methods in NLP often rely on majority or fixed-weight voting,
% which cannot fully exploit performance discrepancies among individual classifiers.
% Recent work has shifted toward learned weighting schemes in which
% a meta-learner assigns weights that maximize metrics such as F1.
% Notably, the RLAE framework~\ref{??} casts the weighting problems as 

\section{RV-HATE}
\label{sec:method}

RV-HATE consists of four modules ($\mathtt{M_0}$--$\mathtt{M_3}$) designed to capture
dataset-specific characteristics.
The three modules~$\mathtt{M_1}, \mathtt{M_2}, \mathtt{M_3}$ extend $\mathtt{M_0}$ by incorporating their specialized functionality for hate speech detection.
We fine-tune each module using contrastive learning and then 
present a voting mechanism that combines the outputs of all
modules to reflect the diverse characteristics of each dataset.
We further enhance the performance via reinforcement learning~\citep{fu2025rlae, birman2022}
that assigns optimized and dataset-specific weights for these modules
during the voting process.
The overall framework is illustrated in Figure~\ref{fig:overview}.

\subsection{Clustering-based Contrastive Learning~($\mathtt{M_0}$)}\label{subsec:module1}
We adopt the SharedCon approach of \citet{AhnKKH24} and refine the model
by modifying the criterion for anchor selection
based on the intuition that
cosine similarity better reflects semantic alignment
in high-dimensional embedding space.
Accordingly, we use cosine similarity instead of Euclidean distance
when measuring the margin between cluster centers and their closest representatives.
Unlike Euclidean distance, which captures absolute magnitude difference, 
cosine similarity focuses on vector direction. This helps to identify 
semantically similar data in contrastive learning better.
The example sentences of cosine similarity and Euclidean distance are shown in Appendix~\ref{app:cos}.

\subsection{Adding [TARGET] Tokens ($\mathtt{M_1}$)}
\label{subsec:module2}

Hate speech is generally defined as expressions conveying hatred or discrimination toward a specific target, and it is considered a subset of abusive text~\citep{poletto2021}.
Abusive (offensive) text refers to any expression intended to insult,
humiliate, threaten, or harass another person.
The primary distinction between hate speech and abusive text lies
in the presence of a \textbf{specific target}.
% Therefore, if a sentence expresses insults without clear targets, it is abusive text, but cannot be considered hate speech.
Therefore, a sentence that expresses insults without clear targets is considered abusive text, not hate speech.
However, we observe that this boundary is often blurred
in widely used hate speech datasets (Appendix~\ref{sec:offensive/hate})
where offensive and hateful language are often hard to distinguish.
We therefore design $\mathtt{M_1}$ that tags tokens referring to specific groups or
institutions in hate-labeled data to address this issue. 
We use spaCy~\cite{Honnibal17} for Named Entity Recognition (NER) tagging
and gpt-4o as a supplementary tagger when spaCy fails to tag certain entities.
Tagging with spaCy covers approximately 18.75\% of the data on average,
and supplementary tagging with gpt-4o increases this coverage to almost 50.88\%.
Following the previous study~\cite{KhuranaNF25}, we specifically focused on [ORG] (organization), [NORP] (nationalities), and [GPE] (country) entities as target tokens for RV-HATE.
% By concatenating this tagged hate-labeled data with the original training dataset,
We augment the training dataset with target-tagged hate-labeled data,
allowing the model to better understand context through target entities and
distinguish between offensive language and hate speech.

\subsection{Outlier Removal within Clusters ($\mathtt{M_2}$)}
\label{subsec:module3}

Hate speech datasets often contain broken sentences, 
because they are primarily collected through web crawling.
A broken sentence refers to an incomplete or fragmented sentence 
that lacks essential grammatical components.
We analyze 500 randomly sampled data from each dataset using gpt-4.1
to quantify the presence of such sentences.
The analysis reveals that all datasets contain broken sentences, with an average proportion of 30.28\%.
Previous studies have shown that broken sentences hinder
semantic understanding and may interfere with model learning~\cite{AhnKKH24}. 
In our analysis, we observe that broken sentences often exhibit
abnormal representations in embedding space and tend to lie
farther from the cluster center.
Examples of broken sentences are provided in Appendix~\ref{app:broken_sentence}.
This suggests that they behave similarly to outliers in clusters.

During training, RV-HATE computes the center of each cluster to select anchors.
% However, clusters often include outliers
% with many broken sentences appearing as such outliers.
However, if clusters contain broken sentences (\textit{e.g.}, those with typos or incomplete masking),
they may act as outliers and degrade the quality of anchor selection.
% Examples of broken sentences are provided in Appendix~\ref{app:broken_sentence}.
% These outliers distort anchor computation and reduce its accuracy.
To address this issue, $\mathtt{M_2}$ applies the InterQuartile Range (IQR) method, 
a well-established statistical approach for outlier detection~\citep{mramba2024detecting}.
This method computes the upper-bound threshold 
based on the IQR of the distances from the cluster center (Appendix~\ref{app:iqr}).
Data points exceeding this threshold are removed, and
the cluster center is recalculated.
This outlier removal process reduces the influence of broken sentences 
and improves the overall quality of the datasets.
The proportion of removed outliers appears in Appendix~\ref{app:outlier_remove}.

\subsection{Using Hard Negative Samples ($\mathtt{M_3}$)}
\label{subsec:module4}

Hate speech datasets inherently contain a high level of noise
due to the subjective nature of human annotation.
Annotators often interpret hate speech differently because of personal bias, 
background knowledge or contextual perception~\cite{KhuranaNF25}.

We randomly select 500 samples from each dataset
and use gpt-4.1 to quantify the proportion of mislabeled instances.
On average, 18.24\% of the data across datasets is identified as mislabeled (Appendix~\ref{app:gpt-mislabeled}).
Such labeling inconsistencies negatively affect model training,
by introducing ambiguous decision boundaries and decreasing classification performance~\cite{AhnKKH24}.
Therefore, we propose $\mathtt{M_3}$ to identify and leverage hard negatives near the decision boundary
for better discrimination.
Hard negative samples include data points with high cosine similarity to the anchor
yet belong to a different class.
Additionally, false positive samples with high model confidence serve as hard negatives.
Unlike standard contrastive learning, where only in-batch negative samples are selected,
we employ a queue to store hard negative samples from multiple batches~\citep{KimJPPH24}.
This allows the model to capture challenging negatives beyond the current batch
and extends the selection process across a broader range of data.
Consequently, the model learns more refined decision boundaries and enhances classification performance.

\subsection{Reinforcement Learning-Guided Soft Voting}
\label{subsec:voting}

We propose a reinforcement learning-guided soft voting mechanism for effective detection of implicit hate speech across datasets.
% with diverse linguistic and contextual characteristics.
This approach enables an adaptive ensemble strategy by dynamically learning the weight
to assign to each module depending on dataset-specific characteristics.
We independently train the four classifiers $f_k$ ($k\in \{0, 1, 2, 3\}$) based on each module designed to capture complementary aspects of implicit hate speech.
Each classifier $f_k$ outputs a logit vector 
\[
z_{k,i}=[z_{k,i}^{(0)}, z_{k,i}^{(1)}],
\]
where $i$ indexes an input and $(0), (1)$ denote non-hate and hate
classes, respectively.

\paragraph{Soft Voting.}
We aggregate the predictions of the four classifiers by computing a weighted average of their logits.
The ensemble logit for each class $h\in \{0,1\}$ is given by
\begin{equation*}
    Z_i^{(h)}=\sum_{k=0}^3 w_k\cdot z_{k,i}^{(h)},
\end{equation*}
where $w_k$ denotes the reinforcement learning-optimized vector (module-weights). 
% These weights satisfy the following constraints:
% \begin{equation}
%     \sum_{k=1}^{4}w_k = 1, \quad w_k\ge 0.
% \end{equation}
We predict the final label by selecting the class with the highest ensemble logit.
\begin{equation*}
    \hat y_i=\arg\max_{h\in\{0,1\}}Z_i^{(h)}.
\end{equation*}
This approach enables the ensemble to focus on models that perform more
reliably under the specific characteristics of the input hate speech data.

\paragraph{Reinforcement Learning.}

% The model weights are dynamically optimized via reinforcement learning.
We formulate the model weights assignment problem as a sequential decision-making task,
where a policy network $\pi_\theta(\textbf{w}|s)$ generates the weight vector $\textbf{w}=[w_0, w_1, w_2, w_3]$ conditioned on the current state $s$.
After sampling $\textbf{w}$,
we apply soft voting and evaluate the resulting prediction $\hat y$ on a validation set.
The F1-based reward $r$ guides policy optimization.
We adopt Proximal Policy Optimization (PPO)~\cite{Schulman17} to train the policy
network by maximizing the following objective:
\begin{equation*}
r_t(\theta)=\frac{\pi_\theta(\textbf{w}_t|s_t)}{\pi_{\theta_{old}}(\textbf{w}_t|s_t)},
\end{equation*}
\begin{equation*}
L(\theta)=\mathbb{E}_t[\min(r_t(\theta)\hat{A}_t, \mathrm{clip}(r_t(\theta), 1-\epsilon, 1+\epsilon)\hat{A}_t)],
\end{equation*}
where $r_t(\theta)$ is the probability ratio between the new and old policies at timestep $t$,
and $\hat{A}_t$ is the estimated advantage, measuring how much better the selected
action $\textbf{w}_t$ performs than a baseline.
The expectation $\mathbb{E}_t$ is taken over timesteps in a batch of episodes.
Each timestep corresponds to a forward pass 
where the policy samples a weight vector $\textbf{w}_t$ and receives a reward.
PPO optimizes $L(\theta)$ over these steps, with clipping applied to the probability ratio $r_t(\theta)$ to prevent large policy updates when $r_t(\theta)$ deviates from 1, ensuring stable convergence.

% clipping term preventing large
% policy updates when $L(\theta)$ deviates from 1.0, ensuring stable convergence. 
$\epsilon$ is a clipping parameter that limits the policy update step.
We initialize all weights $w_k=0.25$ and constrain them during training
to remain positive and sum to 1, while the policy network learns dataset-specific weight configurations that maximize ensemble performance.

% Our method learns to assign model weights based on the data 
% which allows the ensemble to emphasize more reliable classifiers for each dataset.
% This approach improves the detection performance of subtle 
% and context-dependent implicit hate speech.

\begin{table*}[h!]
\centering
    \begin{tabular}{l|ccccc|c}
    \noalign{\hrule height 0.8pt}
     & \multicolumn{5}{c|}{\small{\textbf{Datasets}}} & \multirow{2}{*}{Average} \\ 
    \makecell{\small{\textbf{Models}}} & IHC & SBIC & DYNA & Hateval & Toxigen & \\ 
    \noalign{\hrule height 0.8pt \vskip 2pt}
    CE & 77.70 & 83.80 & 78.80 & 81.11 & 90.06 & 82.29\\
    SCL \citep{KhoslaTWSTIMLK20} & 77.81 & 82.92 & 80.39 & 81.28 & 90.75 & 82.63\\
    % ImpCon \citep{KimPH22} & 78.00 & 83.60 & - & - & - & \\
    SharedCon \citep{AhnKKH24} & 78.50 & 84.30 & 79.10 & 80.24 & 91.21 & 82.67\\
    LAHN \citep{KimJPPH24} & 78.40 & 83.98 & 79.64 & 80.42 & 90.42 & 82.57 \\
    \cdashline{1-7}
    \noalign{\vskip 3pt}
    % RV-HATE (\textit{SharedCon}) & 78.90 \small{$\pm$0.35} & 82.95 \small{$\pm$0.25} & 81.64 \small{$\pm$0.47} & 83.19 \small{$\pm$0.49} & 93.36 \small{$\pm$0.28} & 84.01 \small{$\pm$0.37} \\
    RV-HATE (\textit{Ours}) & \textbf{79.07} & \textbf{84.62} & \textbf{81.82} & \textbf{83.44} & \textbf{93.41} & \textbf{84.47} \\
    \noalign{\hrule height 0.8pt}
    \end{tabular}
\caption{Performance comparison with four baseline methods.
We report the macro-F1 scores averaged over three runs
with different random seeds. 
The \textbf{bold} text indicates the best performance.
RV-HATE shows the best performance across all datasets.}
\label{tab:ID}
\end{table*}

\begin{table*}[ht!]
    \centering\resizebox{\textwidth}{!}{
    \begin{tabular}{l|ccccc|c}
    \noalign{\hrule height 0.8pt \vskip 2pt}
         & IHC & SBIC & DYNA & Hateval & Toxigen & Average \\
    \noalign{\hrule height 0.8pt \vskip 2pt}
        RV-HATE (\textit{combined modules}) & 77.32 \small{$\pm$0.51} & 81.31 \small{$\pm$1.26} & 76.50 \small{$\pm$4.95} & 81.26 \small{$\pm$0.86} & 92.02 \small{$\pm$0.62} & 81.64 \small{$\pm$1.64} \\
        RV-HATE (\textit{equal weights}) & 78.58 \small{$\pm$0.58} & 84.06 \small{$\pm$0.10} & 81.07 \small{$\pm$0.29} & 82.52 \small{$\pm$0.25} & 92.69 \small{$\pm$0.44} & 83.78 \small{$\pm$0.33} \\
        % V-HATE w/o module & 78.56 & 84.61 & 81.55 & 82.87 & 91.93 \\
        RV-HATE ($\ell_2$) & 78.90 \small{$\pm$0.35} & 82.95 \small{$\pm$0.25} & 81.64 \small{$\pm$0.47} & 83.19 \small{$\pm$0.49} & 93.36 \small{$\pm$0.28} & 84.01 \small{$\pm$0.37} \\
        \textbf{RV-HATE (\textit{ours})} & \textbf{79.07} \small{$\pm$0.15} & \textbf{84.62} \small{$\pm$0.23} & \textbf{81.82} \small{$\pm$0.22} & \textbf{83.44} \small{$\pm$0.10} & \textbf{93.41} \small{$\pm$0.21} & \textbf{84.47} \small{$\pm$0.18} \\
    \noalign{\hrule height 0.8pt}
    \end{tabular}}
    \caption{Performance comparison of RV-HATE and its variants.
    The \textit{combined modules} (Sec.~\ref{subsec: integration-module}) refer to a single model
    trained using all modules without any voting mechanism,
    \textit{equal weights} (Sec.~\ref{subsec:rv}) indicates the voting performance 
    when each module is assigned an equal weight of 0.25,
    and $\ell_2$ denotes the base model using the Euclidean distance instead of cosine similarity (Sec.~\ref{subsec:cosine}).
    %Single model with integrated four modules, RV-HATE with fixed weights and RV-HATE with reinforce weights across different datasets. 
    We report the macro-F1 scores averaged over three runs with different random seeds. The \textbf{bold} text indicates the best performance. 
    % RV-HATE achieves the best outcomes across all datasets. 
   }
    \label{tab:pitfalls}
\end{table*}

\section{Experimental Results}

\subsection{Datasets}
We conduct experiments on five hate speech datasets---IHC, SBIC, DYNA, Hateval, and Toxigen---that 
cover a broad spectrum of characteristics for a comprehensive evaluation.
The detailed settings and dataset explanations are provided in Appendix~\ref{app:used_dataset}, \ref{app:data-statistics}.

\subsection{Baselines}
We compare our approach with four baseline methods:
\textbf{CE} is a general approach for hate speech detection based on cross-entropy loss.
\textbf{SCL}~\citep{KhoslaTWSTIMLK20} is a supervised contrastive learning that uses labels to bring representations of the same class closer and push apart representations of different classes.
% \item \textbf{ImpCon} utilizes an implication as an anchor in the process of contrastive learning. 
% ImpCon works well in implicit hate speech detection when explicit implications are present in the datasets, but does not work without implication texts.
\textbf{SharedCon}~\citep{AhnKKH24} is the current SOTA method in implicit hate speech detection.
This method uses the data closest to the center of each cluster as its anchor instead of explicit implications. 
% SharedCon overcomes the limits of Impcon. 
% We use SharedCon as the base model $\mathtt{M_0}$.
\textbf{LAHN}~\citep{KimJPPH24} uses hard negative samples in contrastive learning. 
Hard negatives are data samples that are close to an anchor but have different labels. 
LAHN illustrates the importance of hard negative samples.

\subsection{Implementation Details}

For our experiments, we use a pre-trained language model BERT-base-uncased (110M)~\citep{DevlinCLT19} as the base model and Sim-CSE\footnote{princeton-nlp/unsup-simcse-bert-base-uncased}~\citep{GaoYC21} as a text embedding model.
We train the models on each of the five datasets for 6 epochs with NVIDIA RTX 4090.
For hyperparameter setting, we select the learning rate from \{2e-5, 3e-5\}, 
the temperature $\tau$ from \{0.3\}, $\lambda$ from \{0.5, 0.75\}, 
the number of clusters from \{20, 75, 125\}.
We conduct 10,000 steps for reinforcement learning.
All experiments are executed with three different random seeds.
We report the average score of macro-F1 because
it is more appropriate for evaluating performance on imbalanced hate speech datasets.

\subsection{Experimental Results}

%We evaluate the performance of our approach 
%across five hate speech datasets.

On the Hateval dataset (Table~\ref{tab:ID}), contrastive learning methods 
such as SharedCon and LAHN underperform compared to the 
cross-entropy (CE) baseline (81.11\%).
In contrast, RV-HATE achieves 83.44\%, outperforming the CE baseline by 2.33\%p.
On Toxigen, RV-HATE outperforms SharedCon by 2.2\%p. 
RV-HATE achieves SOTA performance across diverse conditions,
outperforming the prior leading model, SharedCon, by an average of 1.8\%.
% Although a 1--2\% gain may appear minor, 
% It represents a meaningful advance in hate speech detection,
% a task in which model performance trends to plateau around 80\%.
% Indeed, prior studies have also regarded improvements of this magnitude as
% significant progress in the field~\cite{AhnKKH24, KimJPPH24}.
These results demonstrate that the proposed modules and reinforcement learning-based weighting effectively address dataset-specific characteristics.
Section~\ref{subsec:ablation} explains in detail how each module contributes to performance improvements on individual datasets
and Appendix~\ref{app:weights} reports the weight values assigned to each module and dataset.

\label{sec:analysis}
\begin{table*}
    \centering
    \begin{tabular}{c|ccccc|c}
    % \noalign{\hrule height 0.8pt \vskip 2pt}
    \toprule
     & IHC & SBIC & DYNA & Hateval & Toxigen & Average \\
    % \noalign{\hrule height 0.8pt \vskip 2pt}
    \toprule
    $\mathtt{M_0}$ & 77.26 & 83.36 & 80.52 & 81.02 & 91.25 & 82.68 \\
    $\mathtt{M_1}$ & 77.53 & 82.94 & 79.51 & 81.63 & 90.55 & 82.43 \\
    $\mathtt{M_2}$ & 77.64 & 83.11 & 80.26 & 81.45 & 92.01 & 82.89 \\
    $\mathtt{M_3}$ & 77.42 & 83.28 & 79.87 & 81.78 & 92.63 & 83.00 \\
    % \noalign{\vskip 2pt}
    \midrule
    % \noalign{\vskip 3pt}
    RV-HATE & \textbf{79.07} & \textbf{84.62} & \textbf{81.82} & \textbf{83.44} & \textbf{93.41} & \textbf{84.47}  \\
    \hspace{10pt} ~~~- $\mathtt{M_0}$ & 79.04 \small(-0.03) &  84.28 \small(-0.34) &  81.15 \small(-0.67) & 83.20 \small(-0.24) & 93.16 \small(-0.25) & 84.17 \\
    \hspace{10pt} ~~~- $\mathtt{M_1}$ & 78.37 \small(-0.70) &  84.50 \small(-0.12) &  81.56 \small(-0.26) & 83.04 \small(-0.40) & 92.99 \small(-0.42) & 84.09 \\
    \hspace{10pt} ~~~- $\mathtt{M_2}$ &  78.60 \small(-0.47) &  84.36 \small(-0.26) &  81.66 \small(-0.16) &  83.01 \small(-0.43) &  93.16 \small(-0.25) & 84.15 \\
    \hspace{10pt} ~~~- $\mathtt{M_3}$ &  78.79 \small(-0.28) &  84.24 \small(-0.38) &  81.17 \small(-0.65) &  82.88 \small(-0.56) &  92.87 \small(-0.54) & 83.99 \\
    % \noalign{\hrule height 0.8pt}
    \bottomrule
    \end{tabular}
    \caption{Ablation study results for RV-HATE. The table shows the performance of each module and the impact of excluding each module. Each result represents the average macro-F1 score of three runs with different random seeds. The value in parentheses indicates the performance change compared to RV-HATE, and \textbf{bold} text indicates the best performance.
    The model achieves strong performance when utilizing all modules jointly.}
    \label{tab:ablation}
\end{table*}

\begin{table*}[t!]
    \centering{
    \begin{tabular}{c|ccccc}
    \noalign{\hrule height 0.8pt \vskip 2pt}
        Type & IHC & SBIC & DYNA & Hateval & Toxigen \\
    \noalign{\hrule height 0.8pt \vskip 2pt}
        entity-tagged~($\mathtt{M_1})$ & 67.83 & 65.04 & 27.56 & 48.44 & 48.53 \\
        outlier-removed~($\mathtt{M_2})$ & 0.59 & 0.44 & 0.38 & 0.69 & 0.31 \\
        % Mislabeled data & $\triangle$ & $\bigcirc$ & $\times$ & $\bigcirc$ & $\times$ \\
    \noalign{\hrule height 0.8pt}
    \end{tabular}}
    \caption{
    Dataset statistics for $\mathtt{M_1}$ and $\mathtt{M_2}$.
    The table shows the ratio of entity-tagged data and outlier-removed data in each dataset after applying the $\mathtt{M_1}$, $\mathtt{M_2}$ modules.
    }
    \label{tab:datasets_features}
\end{table*}

\section{Analysis}

\subsection{The Integration of all Modules}
\label{subsec: integration-module}

We train a single model that integrates all four modules
($\mathtt{M_0}$, $\mathtt{M_1}$, $\mathtt{M_2}$ and $\mathtt{M_3}$)
to examine the impact of the voting mechanism, which is called `combined modules'.
As shown in Table~\ref{tab:pitfalls}, RV-HATE (\textit{combined modules}) consistently exhibit worse performance compared to RV-HATE (\textit{ours})---average decrease of 2.83\%p.
% Specifically, the performance on the Toxigen dataset is 2.96\%p lower 
% and the other datasets show an average decrease of 2.09\%p.
This performance drop indicates that jointly training on all modules leads
to a loss of specialization, reducing the ability of the model to adapt to data-specific characteristics.
In contrast, RV-HATE (\textit{ours}) retains its specialization,
allowing the voting mechanism to leverage diverse perspectives.
% Examples of inference results of each module are in Appendix~\ref{??}
These results highlight that preserving modular specialization
and leveraging their complementary views is more effective than
combining them into a unified model.

\subsection{The Use of Reinforced Voting Mechanism}
\label{subsec:rv}
We analyze the impact of reinforcement learning on the voting mechanism in RV-HATE.
We evaluate its effectiveness by conducting an experiment 
using fixed weights $[0.25, 0.25, 0.25, 0.25]$ without reinforcement learning-based weights (RV-HATE (\textit{equal weights})).
In contrast, RV-HATE (\textit{ours}) learns dataset-specific optimal weights through reinforcement learning,
enabling the voting mechanism to reflect the contribution of each module for a given dataset.
When comparing the two settings in Table~\ref{tab:pitfalls}, 
we observe that the approach using optimized weights (RV-HATE (\textit{ours}))
achieves an average improvement of 0.68\%p.
The weights are optimized according to
the characteristics of each dataset,
presented in Appendix~\ref{app:weights}.
This result demonstrates that reinforcement learning effectively identifies the optimal combination of module contributions optimized to each dataset.
However, simple voting with equal weights fails to capture dataset-specific features.
% Since individual modules specialize in different aspects of the data,
% their relative importance varies across datasets.
By adaptively balancing the contributions of specialized modules, reinforcement learning enables the voting mechanism to
leverage dataset-specific expertise and achieve improved performance.

\subsection{The Impact of Cosine Similarity}
\label{subsec:cosine}
We conduct an experiment to examine the impact of using
cosine similarity in module training.
Specifically, we compare RV-HATE (\textit{ours}) that uses cosine
similarity for the modules and 
RV-HATE (\textit{$\ell_2$}) that uses Euclidean distance.
As shown in Table~\ref{tab:pitfalls},
RV-HATE (\textit{ours}) achieves an average improvement of 0.46\%p
and shows a lower standard deviation across datasets.
These results validate the effectiveness of employing 
cosine similarity over Euclidean distance for training the modules.

\begin{table*}[t!]
    \centering{
    \begin{tabular}{c|ccc}
    \noalign{\hrule height 0.8pt \vskip 2pt}
        Model & Number of Parameters & Training Time & Inference Latency \\
    \noalign{\hrule height 0.8pt \vskip 2pt}
        SharedCon & 110M & 1h & 0.5-1.5ms \\
        LAHN & 110M & 1h & 0.5-1.5ms \\
        RV-HATE & 110M * 4 + 4,085 (PPO) & 1h * 4 + 5-10m (PPO) & 0.5-1.5ms * 4 \\
    \noalign{\hrule height 0.8pt}
    \end{tabular}}
    \caption{
    Comparison of model complexity, training time, and inference latency for RV-HATE and encoder-based baselines. All modules use BERT-base encoders. Training time is measured on the IHC dataset.
    }
    \label{tab:efficiency}
\end{table*}

\subsection{Ablation Study}
\label{subsec:ablation}
% Table~\ref{tab:ablation} presents the effect of each module on overall performance,
% including both standalone results and ablation study outcomes.
We analyze the performance of RV-HATE and its individual modules.
% $\mathtt{M_0}$ serves as the baseline module utilizing clustering-based contrastive learning,
% while $\mathtt{M_1}$ to $\mathtt{M_3}$ build upon $\mathtt{M_0}$ by integrating additional functionalities.
Table~\ref{tab:ablation} presents the performances of each module.
Notably, some modules perform worse than the baseline ($\mathtt{M_0}$).
This suggests that individual modules may be biased as they consider
dataset-specific characteristics.
In contrast, RV-HATE achieves a higher performance than $\mathtt{M_0}$ alone,
as the voting mechanism mitigates such biases and balances the variance across modules.
By incorporating diverse dataset-specific characteristics, 
RV-HATE can achieve enhanced overall performance.

\paragraph{Module 1 (\textit{$\mathtt{M_1}$}).}
% addresses the hypothesis that existing hate speech datasets 
% often lack a clear boundary between offensive and hate speech.
% As shown in Table~\ref{tab:ablation}, 
Excluding $\mathtt{M_1}$ from IHC causes the largest
performance drop (-0.7\%p).
Table~\ref{tab:datasets_features} supports this finding by showing that 
IHC contains the highest proportion of NER-tagged data.
It indicates that implicit hate speech often relies on subtle target
references that help distinguish it from merely offensive content.
SBIC also has a relatively high proportion of tagged data.
However, since SBIC is an offensive dataset,
excluding $\mathtt{M_1}$ results in only a marginal performance drop (-0.12\%p).
These results support our hypothesis by demonstrating that datasets with
ambiguous boundaries between offensive and hate speech rely on $\mathtt{M_1}$
to capture target-specific cues.
$\mathtt{M_1}$ effectively distinguishes offensive expressions from genuine hate speech and improves the overall performance.
% In order to further investigate this, we randomly sample 500 examples from each dataset
% and use GPT-4.1 to estimate the proportion of implicit hate speech.
Additionally, we analyze the proportion of implicit hate speech in each dataset. 
As shown in Figure~\ref{fig:ablation_ratio} of Appendix~\ref{app:ablation-analysis}, IHC exhibits the highest proportion of implicit hate speech data,
which aligns with its reliance on $\mathtt{M_1}$.
% as reflected in Table~\ref{tab:ablation}.

\paragraph{Module 2 ($\mathtt{M_2}$).}
% is a module introduced based on the hypothesis that
% typos, masking and other broken sentences often constitute outliers
% within a cluster.
% As shown in Table~\ref{tab:ablation}, e
Excluding $\mathtt{M_2}$ from IHC and Hateval results in a performance drop of 0.47\%p, 0.43\%p, respectively.
Consistently, Table~\ref{tab:datasets_features} shows that 
the IHC and Hateval have the highest proportions of removed data.
This performance decline indicates that $\mathtt{M_2}$ plays a crucial role in mitigating
the impact of outliers.
These findings demonstrate the effectiveness of $\mathtt{M_2}$ in handling noisy data
and improving detection in datasets with high levels of textual noise.
% This pattern is echoed in Figure~\ref{fig:ratio:broken}. 
% In order to quantify the prevalence
% of broken sentences, 500 examples are randomly sampled from each dataset and we use
% GPT-4.1 for analysis. 
Additionally, we analyze the distribution of broken sentences.
As shown in Figure~\ref{fig:ablation_ratio} of Appendix~\ref{app:ablation-analysis}, IHC exhibits the highest proportion of broken sentences,
whereas DYNA shows the lowest.

\paragraph{Module 3 ($\mathtt{M_3}$).}
% is designed based on the hypothesis that
% distinguishing hard negatives near the decision boundary can improve
% performance.
Removing $\mathtt{M_3}$ from RV-HATE has the greatest impact across
all datasets.
Moreover, the standalone averaged performance of the $\mathtt{M_3}$ module
surpasses that of the other modules.
We examined the embedding representations of each dataset using
t-SNE (Figure~\ref{fig:app_embeddings} in Appendix~\ref{app:embeddings}) to validate whether $\mathtt{M_3}$
indeed clarifies the decision boundary.
The visualizations show that the decision boundary becomes more
clearly separated compared to the baseline module.
More detailed results are presented in Appendix~\ref{app:embeddings}.
These findings confirm that $\mathtt{M_3}$ effectively enhances the separability
of hard negatives near the decision boundary, 
thereby improving the ability to detect subtle and ambiguous instances.

% \begin{table*}[h!]
% \centering
% \begin{tabular}{lccp{8.3cm}l}
% \toprule
% Dataset & Error Type & Ratio & Example & label \\
% \midrule
% \multirow{3}{*}{IHC}     & $\mathtt{type1}$ & 23.85\% & ``this is so unislamic -- no compulsion in religion'' & non-hate \\
%         & $\mathtt{type2}$    & 45.15\% & whine all you want davey. white id politics is here to stay. & non-hate \\
%         & $\mathtt{type3}$     & 24.80\% & explicitly huwhite & non-hate \\
%         & $\mathtt{type4}$     & 6.20\% &  & n/a \\
% \midrule
% \multirow{3}{*}{SBIC}    & $\mathtt{type1}$     & 15.90\% & \# health... yo-ho-ho and a bottle of \# rum! & non-hate \\
%         & $\mathtt{type2}$     & 33.38\% & any mens rights groups / people here in sf? & hate \\
%         & $\mathtt{type3}$     & 23.35\% & these days bitches be just letting anyone hit raw!! & not-hate \\
%         & $\mathtt{type4}$     & 27.36\% &  &  n/a\\
% \midrule
% \end{tabular}
% \caption{Error analysis of RV-HATE, where 
% $\mathtt{type1}$ is the case of broken sentences, $\mathtt{type2}$ is the case of  ambiguous sentences
% that are difficult to classify even by human experts,
% $\mathtt{type3}$ is the case of clearly mislabeled,
% $\mathtt{type4}$ is the case of the rest.
% The `label' column shows the ground-truth label.
% }
% \label{tab:error_types1}
% \end{table*}

\begin{table*}[h!]
\centering
\begin{tabular}{lccp{8.3cm}l}
\toprule
Dataset & Error Type & Ratio & Example & label \\
\midrule
\multirow{3}{*}{IHC}     & $\mathtt{type1}$ & 23.85\% & ``this is so unislamic -- no compulsion in religion'' & non-hate \\
        & $\mathtt{type2}$    & 45.15\% & whine all you want davey. white id politics is here to stay. & non-hate \\
        & $\mathtt{type3}$     & 24.80\% & explicitly huwhite & non-hate \\
        & $\mathtt{type4}$     & 6.20\% &  & n/a \\
\midrule
\multirow{3}{*}{SBIC}    & $\mathtt{type1}$     & 15.90\% & \# health... yo-ho-ho and a bottle of \# rum! & non-hate \\
        & $\mathtt{type2}$     & 33.38\% & any mens rights groups / people here in sf? & hate \\
        & $\mathtt{type3}$     & 23.35\% & these days bitches be just letting anyone hit raw!! & not-hate \\
        & $\mathtt{type4}$     & 27.36\% &  &  n/a\\
\midrule
\multirow{4}{*}{DYNA}    & $\mathtt{type1}$     & 5.31\% & wooooooooooman = trrassssshhh & hate \\
        & $\mathtt{type2}$     & 33.20\% & i see no difference between toffs and entitled children & not-hate \\
        & $\mathtt{type3}$     & 44.76\% & men are such a fucking stupid knob heads! & not-hate \\
        & $\mathtt{type4}$     & 16.73\% &  & n/a \\
\midrule
\multirow{4}{*}{Hateval} & $\mathtt{type 1}$ & 25.35\% & correction - hugh if true. \# makemeasandwich & hate \\
 & $\mathtt{type 2}$ & 27.70\% & when you join a relationship women leave this part out \# yesallmen & hate \\
 & $\mathtt{type 3}$ & 41.78\% & u a bitch & hate \\
 & $\mathtt{type 4}$ & 5.16\% & & n/a \\
\midrule
% \multirow{4}{*}{Toxigen} & $\mathtt{type 1}$ & 0\% & - & hate\\
 \multirow{3}{*}{Toxigen} & $\mathtt{type 2}$ & 50.00\% & go back to china. & hate\\
 & $\mathtt{type 3}$ & 21.05\% & ``it is not ok for me to be homophobic, but it is ok for me to be homophobic'' & hate\\
 & $\mathtt{type 4}$ & 28.95\% & & n/a \\
\bottomrule
\end{tabular}
\caption{Error analysis of RV-HATE, where 
$\mathtt{type1}$ is the case of broken sentences, $\mathtt{type2}$ is the case of  ambiguous sentences
that are difficult to classify even by human experts,
$\mathtt{type3}$ is the case of clearly mislabeled,
$\mathtt{type4}$ is the case of the rest.
The `label' column shows the ground-truth label.
Note that there is no $\mathtt{type1}$ error in Toxigen.
}
\label{tab:error_types1}
\end{table*}

\subsection{Computational Efficiency}

We analyze the computational cost of RV-HATE
and compare it with existing encoder-based baselines in Table~\ref{tab:efficiency}.
All modules use BERT-base encoders with 100M parameters.
RV-HATE includes four modules, and each module is trained independently.
This allows parallel training across multiple GPUs and keeps the training time comparable to the baseline.

The reinforcement learning component introduces only a small additional cost.
The total number of parameters for PPO-based reinforcement learning is 4,805, which is negligible compared to the encoder models.
The reinforcement learning stage is applied once during training and
takes about 5 to 10 minutes. 
Detailed parameter configurations are provided in Appendix~\ref{app:ppo}.

At inference time, RV-HATE requires four forward passes of BERT-base.
This leads to a linear increase in latency compared to single-encoder models.
However, the method does not use additional decoding or iterative reasoning,
and the computation remains a simple forward-pass ensemble.
RV-HATE increases computation in proportion to the number of modules,
but maintains practical efficiency through parallel training and minimal reinforcement learning overhead.

\section{Error Analysis}

We analyze the false positive or false negative samples from each dataset
to better understand why the RV-HATE fails to detect them correctly.
We categorize these error samples into three different types:
$\mathtt{type1}$ refers to broken instances that contain grammatical errors,
typographical errors or special characters;
$\mathtt{type2}$ denotes ambiguous instances that are difficult to clearly
classify as hate or not-hate;
$\mathtt{type3}$ corresponds to mislabeled instances;
and $\mathtt{type4}$ includes all remaining cases that cannot be assigned to any of the aforementioned categories.
Two experts annotate the error samples according to these three types.

Table~\ref{tab:error_types1} reports the distribution of error types
for each dataset.
IHC shows a high proportion of $\mathtt{type2}$ errors (45.15\%), as the dataset consists of implicit and ambiguous instances.
Error samples from IHC also exhibit a high proportion of ambiguous instances.
SBIC exhibits a comparable proportion of $\mathtt{type2}$ (33.38\%)
and $\mathtt{type3}$ errors (23.35\%).
Since SBIC is an offensive language dataset,
offensive expressions differ in definitions from hate speech or are subject to annotator bias,
resulting in both ambiguous $\mathtt{type2}$ and mislabeled $\mathtt{type3}$  errors.
In the case of DYNA, the majority of errors belong to $\mathtt{type2}$ (33.20\%) and
$\mathtt{type3}$ (44.76\%), with $\mathtt{type3}$ accounting for the largest proportion.
This result suggests that annotation inconsistencies play a significant role in DYNA errors.
In particular, a high proportion of mislabeled and ambiguous instances indicates that DYNA likely contains a variety of dynamic and context-dependent expressions.
Hateval shows the greatest degree of annotator noise $\mathtt{type3} $(41.78\%), indicating a large number of mislabeled instances.
Manual inspection further confirms that the dataset contains substantial label noise.
In contrast, as a machine-generated dataset, Toxigen exhibits no typographical errors or broken sentences.
However, $\mathtt{type2}$ errors account for 50\% of the error samples,
indicating that Toxigen includes many semantically ambiguous instances
despite its clean text. 
Further analyses are provided in Appendix~\ref{app:error_analysis}.
% Toxigen은 machine-generated dataset이기 때문에 오타와 masking이 깨진 것들의 비율이 현저히 낮았다.
% 그럼에도 불구하고, ambiguous한 데이터는 많이 존재했는데, error sample중 무려 50\%나 type2였다.
% 따라서 Toxigen은 오타는 적지만, ambiguous한 데이터가 많이 존재하는 데이터셋임을 알 수 있다.
% Further analyses on other datasets are provided in Appendix~\ref{app:error_analysis}.

This analysis highlights that ambiguous and mislabeled instances constitute a major
source of performance degradation across datasets.
% Addressing issues such as annotation quality, refining definitions and improving
% strategies for handling ambiguity will be an important direction for building more
% robust and reliable hate speech detection models.

\section{Conclusions}
We have proposed RV-HATE, a reinforcement learning-based voting method for implicit 
hate speech detection that can capture the dataset-specific 
characteristics through the multi-module design.
Each module is designed to address a particular aspect of the datasets.
$\mathtt{M_0}$ focuses on improving contextual understanding 
through cosine similarity,
$\mathtt{M_1}$ enhances the detection of implicit hate speech 
by incorporating hate target tagging,
$\mathtt{M_2}$ removes outliers during training to preserve reliable
data samples,
and $\mathtt{M_3}$ provides a clear decision boundary.
Our approach employs reinforcement learning to assign an adaptive
weight to each module.
By analyzing how each module contributes across different datasets, we have a better understanding of their characteristics.
Thus, through the voting strategy, RV-HATE is able to adapt dataset-specific features better and achieves SOTA performance on multiple benchmarks.

\section*{Limitations}
RV-HATE effectively captures the characteristics of implicit hate speech datasets
and can decide the best module combination with the reinforced voting mechanism.
% Yet, there is still an open question in dealing with synthesized or machine-generated
% sentences (\textit{i.e.}, Toxigen).
In our experiments, $\mathtt{M_1}$ the target-tagging module does not
consistently provide the same improvements on machine-generated samples,
potentially due to style and distribution differences.
Although these results do not diminish the overall utility of RV-HATE, they highlight an opportunity to explore more specialized strategies for 
artificially generated data.
We believe that with further adaptation and refinement, 
RV-HATE’s modular design could be extended to manage artificially 
generated text effectively and
broaden its applicability in future work.

\section*{Ethical Consideration}
\paragraph{Minimizing Exposure Risks}
Existing methods rely on annotations that human annotators directly label
and explain the meaning of hateful sentences.
On the other hand, our approach allows the model to learn from 
representative samples without requiring manually annotated implications.
As a result, our method is expected to reduce the mental load
on annotators and contribute to a more ethical data collection process.

\paragraph{Dataset-aware Hate Speech Detection}
Focusing solely on improving model generalization can overlook important
dataset-specific characteristics.
Our approach introduces a voting methodology that employs the unique
hate speech patterns of each dataset to make optimal decisions.
By capturing diverse forms of hate speech while respecting 
the contextual nuances of individual datasets, 
our method contributes to the development of a more reliable 
and context-aware hate speech detection system.

\paragraph{Risks and Potential Misuse}
Our detection capability is designed to identify hate speech effectively; however, there remains a possibility that it could be leveraged in unintended ways, such as generating new forms of hate speech.
Addressing these risks requires careful monitoring of
how the model is used and a critical assessment of its impact.

\section*{Acknowledgments}
This research was supported by the KISA grant (RS-2025-02222626)
and the AI Graduate School Program (RS-2020-II201361)
funded by the Korean government.

\bibliography{custom}

\appendix

\section{Sentence Examples}
\subsection{Examples of Hate without Target}
\label{sec:offensive/hate}
Table~\ref{tab:hate_wo_target} shows some examples of misannotated sentences in hate speech datasets.
While hate speech is generally defined by the presence of an explicit hate target, these examples lack such a target and are appropriately categorized as offensive language rather than hate speech.
Such inconsistencies in annotation may lead to confusion during
model training and potentially degrade classification performance.
\begin{table*}[h!]
    \centering\resizebox{\textwidth}{!}{
    \begin{tabular}{c|l}
    \noalign{\hrule height 0.8pt}
        \textbf{Dataset} & \textbf{Sentence} \\
    \noalign{\hrule height 0.8pt}
        \multirow{4}{*}{IHC} & when you drab the swamp in sacramento! \\
        \cdashline{2-2}
         & according to one person i communicated with-why cry over as bunch of cells, would you cry over a tumor \\
         \cdashline{2-2}
         & one monkey don't stop no show \\
    \cline{1-2}
        \multirow{3}{*}{Hateval} & When it's my birthday but you're the one that's getting spanked \\
        \cdashline{2-2}
        & bitch I was fuckn up till 4:30 but your hoe ass didn't text me back. \\
        \cdashline{2-2}
        & He can choke and so does his bitch hoe \\
    \cline{1-2}
        \multirow{5}{*}{DYNA} & Rape should be legal \\
        \cdashline{2-2}
        & Something tells me this woman doesnt have a stable and successful life. \\
        \cdashline{2-2}
        & our country is broken. It's basically free of opportunity now, it's overrun by animals and we've lost all \\
        & control of our institutions. I don't want to stay but I cant exaxtly leave, \\
    \cline{1-2}
        \multirow{4}{*}{Toxigen} & there is no difference in the way they act, they are all the same, they all act the same way, they are all the \\
        & same \\
        \cdashline{2-2}
        & no matter how much money you give them, they'll never be as good as us and so on and so on and so on \\
        \cdashline{2-2}
        & i dont care if you are in pain or not, i dont want to hear about it \\
    \noalign{\hrule height 0.8pt}
    \end{tabular}}
    \caption{Examples of sentences that are classified as hate speech but are actually offensive language. 
    These samples are drawn from the IHC, Hateval, DYNA, and Toxigen datasets. 
    The SBIC dataset is excluded as it is basically an offensive language dataset.}
    \label{tab:hate_wo_target}
\end{table*}

\subsection{Examples of Broken Sentences}
\label{app:broken_sentence}
A broken sentence refers to a sentence which lacks essential 
grammatical components.
Table~\ref{tab:broken_sentences} represents examples of broken sentences from the SBIC dataset. 
In the SBIC dataset, numerous instances exhibit such issues,
further compounded by the presence of non-standard characters and
special symbols, which collectively hinder accurate interpretation
and analysis.

\begin{table*}[h!]
\vspace{15pt}
    \centering
    \begin{tabular}{c|l}
    \noalign{\hrule height 0.8pt}
        \textbf{Class} & \textbf{Sentence} \\
    \noalign{\hrule height 0.8pt}
        \multirow{3}{*}{\textit{Offensive}} & S/O to this hoe tonight ol hating ass http://t.co/HIt4475tyZ \\
        % \cdashline{2-2}
         & RT @gioisblack: Goodbye you faggot http://t.co/oJKZjkgVXL \\
         % \cdashline{2-2}
         & RT @TyrinFlyGuy: Niggas fufu \&amp; bitches is too \\
    \cline{1-2}
        \multirow{3}{*}{\textit{Not Offensive}} & \detokenize{#redneck tow truck. #johndeere #4455 #america https://t.co/iFInqwVRad} \\
        % \cdashline{2-2}
        & \detokenize{RT @k_ree2: &#8220;@Trentdamanz: These hoes are loyal&#8221; plot twist} \\
        % \cdashline{2-2}
        & \detokenize{Yasss brownies &amp; ice cream &#128523; http://t.co/8qLa3JsglG} \\
    \noalign{\hrule height 0.8pt}
    \end{tabular}
    \caption{Examples of broken sentences from the SBIC dataset, categorized by class. 
    These sentences are often found as cluster outliers, 
    which can negatively impact model performance.
    Mitigating this issue, we employ $\mathtt{M_2}$ to remove such outliers.}
    \label{tab:broken_sentences}
\end{table*}

\subsection{Examples of Mislabeled Samples}
Table~\ref{tab:mislabeled} provides examples of mislabeled instances
across datasets.
These instances clearly constitute hate speech, but they
are incorrectly labeled as not-hate.
Such mislabeled data can negatively impact model training by 
introducing noise into the learning process.

\begin{table*}[h!]
\vspace{15pt}
    \centering\resizebox{\textwidth}{!}{
    \begin{tabular}{c|l|c}
    \noalign{\hrule height 0.8pt}
        \textbf{Class} & \textbf{Sentence} & \textbf{Expected Class} \\
    \noalign{\hrule height 0.8pt}
        \multirow{6}{*}{Hate} & i'm not interested in helping you work your sickness out, you need a shrink and a pile of meds, still probably & \multirow{6}{*}{Not Hate}\\
        & wouldn't help you. get help & \\
    \cdashline{2-2}
       & Hungary's New laws target Muslim migrants "to preserve Christian culture of Europe" - Free Speech Time - & \\
    \cdashline{2-2}
       & Women are perfect. Now repeat that until you believe it. & \\
    \cdashline{2-2}
        & It is a total disgrace when tax-payer dollars can go to promoting such degenerate filth. & \\
    \cdashline{2-2}
        & i dont care if they are here legally or not. & \\
    \cline{1-3}
        \multirow{6}{*}{Not Hate} & Can you finally see it now? Its not American vs American. Never was and still isn't! This is Americans & \multirow{6}{*}{Hate} \\
        & vs Transplant-Sleeper-Cell-Capitalists (''Illegal Fucking Corporations') & \\
    \cdashline{2-2}
        & who love illegal aliens. You all have more blood on your hands. You have betrayed America. & \\
        & \detokenize{#ILLEGALimmigrants} & \\
    \cdashline{2-2}
        & we need 2 open a window of real knowledge for born muslims to re-learn \& convert . at least go secular & \\
    \cdashline{2-2}
        & Hanging out with guys does not make you a hoe & \\
    \noalign{\hrule height 0.8pt}
    \end{tabular}}
    \caption{Examples of mislabeled sentences across different classes, sourced from the IHC, SBIC, Hateval, DYNA and Toxigen datasets.
    These mislabeled instances can confuse the model,
    highlighting the importance of effectively leveraging hard negatives, which we address using M4.}
    \label{tab:mislabeled}
\end{table*}

\subsection{Difference Between Euclidean Distance and Cosine Similarity}\label{app:cos}
The followings are examples from all dataset to illustrate 
the differences between Euclidean distance and cosine similarity in 
measuring sentence relationships.
For each anchor post, we provide the sample with the smallest Euclidean distance 
and the sample with the highest cosine similarity.
The results demonstrate that the sample with the highest cosine 
similarity better preserves semantic information, making it a more 
suitable candidate as an anchor sentence.

\begin{itemize}
    \item \textbf{IHC:} \detokenize{" rt  "" : stop the white genocide in south africa ! ( video ) via #whiteresister "" "" """}
    \begin{itemize}
        \item \textbf{Highest Cosine Similarity 1:} \detokenize{the genocide of white america : 1980-2010 | read more : via,"  "" : the white genocide continues : white race now minority in london via "" "" """" rt  "" : stop the white genocide in south africa ! ( video ) via #whiteresister "" "" """,the genocide of white america : 1980-2010 | read more : via}
        \item \textbf{Smallest Euclidean Distance 1:} "  "" : the white genocide continues : white race now minority in london via "" "" """
    \end{itemize}
    \item \textbf{SBIC:} I'm strongly opposed to having `Harriet Tubman` on the \$20 bill. What use would I have for a piece of paper worth \$12?
    \begin{itemize}
        \item \textbf{Highest Cosine Similarity 2:} @JStac825 @ThatAngryNigga these nig**s must be deaf. When's the last time you heard a Roots song on the radio? Quest don't get bit**es!
        \item \textbf{Smallest Euclidean Distance 2:} Roast beef sandwiches are ruined for me I saw someone eating a roast beef sandwich today and all I could think about was a stretched out vag**al flaps destroyed by the c**k carousel.
    \end{itemize}
    \item \textbf{DYNA:} Those black women look like eggplants
    \begin{itemize}
        \item \textbf{Highest Cosine Similarity 3:} What is being presented here is a photograph of a species of "African-aero-ape". Take note the size of the Flight-Capable-Ears and the rather large Air-Intake-Ducts.
        \item \textbf{Smallest Euclidean Distance 3:} Haha, we'd just watch her whining and screaming in combat. She won't ever make it. It should be left to the men. Women display too much ego nowadays. They think that it is their right to replace men. Never gonna happen.
    \end{itemize}
    \item \textbf{Hateval:} a never go respect a who post nothing but have naked pics on here or showing there ass you go be looked at as a rat h**|Im gonna treat women like a hole
    \begin{itemize}
        \item \textbf{Highest Cosine Similarity 3:} Im gonna treat women like a hole
        \item \textbf{Smallest Euclidean Distance 3:} ban muslims banislam sendthemback to their shitholes closetheborders
    \end{itemize}
    \item \textbf{Toxigen:} "ha i post this picture of a black guy on my facebook commenting on how ugly he was he looked like an ape and the
    \begin{itemize}
        \item \textbf{Highest Cosine Similarity 3:} no matter how much money you make, you'll never be as rich as a white person, and youll never be as smart as a white person
        \item \textbf{Smallest Euclidean Distance 3:} hurr if ur chinese change ur name to a white onen my chinese friends are trying to stick up for their stereotype so hard right
    \end{itemize}
\end{itemize}

\section{Prompts Used for ChatGPT}

\subsection{Prompt for Identifying Hate Speech Without Given Targets}

We view the distinction between \textit{hate speech} (which explicitly targets an individual or group)
and \textit{offensive speech} (which may be abusive but lacks a specific target)
as a key challenge in hate speech detection.
While many datasets define hate speech based on the presence of a clear target,
real-world cases are often more ambiguous. 
Some hateful statements do not explicitly reference a group or individual, 
making it difficult for models to differentiate between hate speech and general offensive language.

Therefore, we use a determination prompt (Figure~\ref{fig:targetOX})
that determines whether a given hate speech sentence includes an explicit target
to analyze the prevalence of ambiguous cases in each dataset.

This analysis provides insight into the limitations of existing models and datasets in handling implicit hate speech.
By identifying cases where models fail to make this distinction,
we highlight the need for methods that better capture the nuances of hateful expression.

\begin{figure*}[h]
\begin{tcolorbox}[colback=gray!10,colframe=black,width=\textwidth]
You are a hate speech target verification GPT.
When given a post, follow the instructions below:

1. The provided posts are guaranteed to be labeled as hate speech.\\
2. Hate speech is a form of abusive language that specifically targets individuals or groups based on characteristics such as race, gender, religion, or ethnicity.\\
3. Your task is to determine whether the post contains a target-a specific individual or group being attacked.\\
4. If the post includes a target (e.g., a group or individual being insulted, threatened, or dehumanized), classify it as "target O".\\
5. If the post is abusive but lacks a specific target, classify it as "target X".\\
6. When in doubt, classify as "target O".\\

\textbf{Example 1}\\
Given post: "Immigrants are ruining this country. They should all go back."\\
Your response: target O\\

\textbf{Example 2}\\
Given post: "This is absolutely disgusting. People like this have no place here."\\
Your response: target X\\

\textbf{Remember:} Your response should consist of only a single phrase: either "target O" or "target X". No additional explanation or extra words are allowed.
\medskip
\hrule
\medskip
Post: \texttt{[SENTENCE]}\\

\end{tcolorbox}
\caption{Prompt used for identifying whether a hate speech post contains an explicit target.}
\label{fig:targetOX}
\end{figure*}

% hsan: broken sentence 주석
% \subsection{Prompt for Identifying Broken Sentences}

% \begin{figure}[h]
% \begin{tcolorbox}[colback=gray!10,colframe=black,width=\textwidth]
% You are a broken sentence detect GPT.
% When given a sentence, follow the instructions below:

% 1. The provided sentences are examples of hate speech.\\
% 2. A broken sentence is an incomplete or fragmented sentence that lacks essential grammatical components.\\
% 3. Determine whether the given sentence is broken sentence or not.\\
% 4. Ouput 1 if borken sentence, and 0 if it's not.\\

% \textbf{Example 1}\\
% Given post: An immigrant wouldn't ask that question if his familiy had been murdered by savages.\\
% Your response: 0\\

% \textbf{Example 2}\\
% Given post: Awwwwww they no Kaep aint rape no bitch!!!!\\
% Your response: 1\\

% \textbf{Remember:} You have to print only 0 or 1. As shown in the example, no additional explanation is necessary in your response.

% \medskip
% \hrule
% \medskip
% Post: \texttt{[SENTENCE]}\\

% \end{tcolorbox}
% \caption{Prompt used for identifying whether a hate speech is broken sentence or not.}
% \label{fig:broken}
% \end{figure}

\subsection{Prompt for Identifying Mislabeled Data}
\label{app:gpt-mislabeled}
Labels in hate speech datasets often contain annotation errors 
due to the subjective nature of the task and inconsistencies in human judgments.
Some samples may be incorrectly labeled as hate speech despite 
lacking explicit hateful intent, while others may be misclassified 
as non-hate despite containing discriminatory content. 
We design a verification prompt (Figure~\ref{fig:mislabel}) that 
assesses whether a given post's label correctly reflects its content
to address this issue.

Identifying mislabeled instances allows us to analyze patterns
in label inconsistencies and provides insights into common annotation biases,
which can inform future dataset construction and model training strategies.
% The impact of data quality on model performance is discussed in Section~\ref{}.

\begin{figure*}[h]
\begin{tcolorbox}[colback=gray!10,colframe=black,width=\textwidth]
You are a hate speech verification GPT.
When given a post and its associated label, follow the instructions below:

1. The provided posts are samples from a hate speech dataset.\\
2. Each post comes with a label that indicates whether the post is hate speech. 1 indicates hate, 0 indicates non-hate.\\
3. Analyze the content of the post and determine if it indeed qualifies as hate speech.\\
4. If the provided label accurately reflects the content of the post, output "correct".\\
5. If the label does not match the content of the post, output "incorrect".\\

\textbf{Example 1}\\
Given post: "I can't stand those immigrants who take our jobs."\\
Label: hate speech\\
Your response: correct\\

\textbf{Example 2}\\
Given post: "I love everyone regardless of their background."\\
Label: hate speech\\
Your response: incorrect\\

\textbf{Remember:} Your response should consist of only a single word, either "correct" or "incorrect". No additional explanation or extra words are allowed.
\medskip
\hrule
\medskip
Post: \texttt{[SENTENCE]}\\
Label: \texttt{[LABEL]}

\end{tcolorbox}
\caption{Prompt for verifying labels of the datasets we used.}
\label{fig:mislabel}
\end{figure*}

\subsection{NER Tagging}
We use a Named Entity Recognition (NER) tagging approach to systematically analyze
the target entities in hate speech.
The objective of this process is to identify words in a given sentence 
that explicitly denote a specific group or organization
targeted by hate speech.
Since implicit hate speech often relies on subtle cues, 
correctly recognizing these entities is crucial for understanding the nature of hateful expressions.
Figure~\ref{fig:prompt_ner} presents the prompt used to guide this tagging process.
A single tag [TARGET] is used throughout.
If no explicit target exists in the sentence, the model returns the sentence without modification.

By applying our NER-based tagging mechanism, we aim to make latent target information explicit.
This allows for a clearer identification of the specific groups or organizations referenced in hate speech.

\begin{figure*}[h!]
\begin{tcolorbox}[colback=gray!10,colframe=black,width=\textwidth]
You are a hate speech token tagging GPT.
When given a sentence, follow the instructions below:

1. The provided sentences are examples of hate speech. \\
2. To qualify as hate speech, the sentence must insult or discriminate against a specific group or organization.\\
3. Identify the words in the sentence that represent the specific group or organization.\\
4. Tag only the key words that determine the hate speech with the appropriate category.\\

\textbf{Example}\\
Given sentence: immigrants wouldn't ask that question if his family was murdered by savages\\
Your response: [TARGET] immigrants wouldn't ask that question if his family was murdered by savages\\

You have to use only the `[TARGET]' tag (if there exist other tags, you replace the tag to `[TARGET]').
As shown in the example, no additional explanation is necessary in your response.
There may be multiple key words representing the hate speech target.
If there is no specific target, output the original sentence as is.

\medskip
\hrule
\medskip
Post: \texttt{[Sentence]}
\end{tcolorbox}
\caption{Prompt for NER tagging}
\label{fig:prompt_ner}
\end{figure*}

\section{Dataset Statistics}\label{app:statistics}

\subsection{Used Dataset}
\label{app:used_dataset}

We used five datasets.
\textbf{IHC}~\citep{ElSheriefZMASCY21} is a benchmark for implicit hate speech detection.
It includes 22,584 tweets labeled with corresponding natural language explanations (\textit{i.e.}, implications).
\textbf{SBIC}~\citep{SapGQJSC20} provides over 150,000 structured annotations
of social media posts that capture the implications of social biases and stereotypes,
including their offensiveness, intent, and targeted groups.
\textbf{DYNA}~\citep{VidgenTWK20} is a hate speech dataset 
created through a human-and-model-in-the-loop process, 
incorporating adversarial perturbations to improve the robustness of hate speech detection models.
\textbf{Hateval}~\citep{BasileBFNPPRS19} contains 13,000 English and 6,600 Spanish tweets
annotated for hate speech targeting immigrants and women.
We use only the English portion in our experiments.
\textbf{Toxigen}~\citep{HartvigsenGPSRK22} is a machine-generated dataset of toxic and benign statements
about 13 minority groups, designed to improve implicit hate speech detection.

\subsection{Dataset Split Overview}
\label{app:data-statistics}

We split the dataset into train, validation, and test sets in an 8:1:1 ratio.
The augmented train set was generated by adding target tags only to hate-labeled data, and it only used
in $\mathtt{M_2}$.

\begin{table*}[h!]
    \centering
    \begin{tabular}{ccccc}
    \noalign{\hrule height 0.8pt}
         \textbf{Dataset} & \textbf{Train set} & \textbf{Augmented Train Set} & \textbf{Valid set} & \textbf{Test set} \\
    \noalign{\hrule height 0.4pt}
        IHC & 14,932 & 18,796 & 1,867 & 1,867 \\
        SBIC & 35,504 & 45,290 & 4,673 & 4,698 \\
        DYNA & 33,004 & 44,427 & 4,125 & 4,126 \\
        Hateval & 10,384 & 13,319 & 1,298 & 1,298 \\
        Toxigen & 5,420 & 6,704 & 678 & 678 \\
    \noalign{\hrule height 0.8pt}
    \end{tabular}
    \caption{The statistical information of five datasets in our experiments.}
    \label{tab:data}
\end{table*}

\subsection{Outlier Removal Ratios}\label{app:outlier_remove}
Table~\ref{tab:outlier} summarizes the number and proportion of samples removed by $\mathtt{M_2}$
in each dataset. Hateval show the highest removal rate, whereas Toxigen has the lowest. 
Although the absolute count of discarded samples is modest,
this outlier-removal step impacts RV-Hate's performance by mitigating the influence of noisy data.

\begin{table*}[h!]
    \centering{
    \begin{tabular}{c|ccccc}
    \noalign{\hrule height 0.8pt \vskip 2pt}
         & IHC & SBIC & DYNA & Hateval & Toxigen \\
    \noalign{\hrule height 0.8pt \vskip 2pt}
        Outlier & 66 & 156 & 127 & 72 & 17 \\
        Total  & 11,199 & 35,504 & 33,004 & 10,384 & 5,420 \\
        Ratio (\%) & 0.59 & 0.44 & 0.38 & 0.69 & 0.31 \\
    \noalign{\hrule height 0.8pt}
    \end{tabular}}
    \caption{The ratio of the removed outlier data}
    \label{tab:outlier}
\end{table*}

\section{Weights}
\label{app:weights}
Table~\ref{tab:weights} gives the average reinforcement learning-based voting weights for modules
$\mathtt{M_0}$--$\mathtt{M_3}$ across three random seeds.
For IHC, although $\mathtt{M_0}$ retains a minor share of the vote, the model leans most heavily
on hard negative sampling, then entity tagging, with the least emphasis on removing outliers.
SBIC, Hateval and DYNA concentrate their weight on $\mathtt{M_0}$, with other modules sharing
the rest of the weight moderately.
On the other hand, Toxigen prioritizes hard negative sampling while almost disregarding entity tagging.

\begin{table}[h!]
    \centering{
    \begin{tabular}{c|ccccc}
    \noalign{\hrule height 0.8pt \vskip 2pt}
         & \textbf{M$_0$} & \textbf{M$_1$} & \textbf{M$_2$} & \textbf{M$_4$} \\
    \noalign{\hrule height 0.8pt \vskip 2pt}
        IHC & 0.191 & 0.258 & 0.167 & 0.357  \\
        SBIC  & 0.357 & 0.252 & 0.210 & 0.181 \\
        DYNA & 0.330 & 0.179 & 0.265 & 0.225 \\
        Hateval & 0.327 & 0.187 & 0.248 & 0.238 \\
        Toxigen & 0.227 & 0.001 & 0.314 & 0.459 \\
    \noalign{\hrule height 0.8pt}
    \end{tabular}}
    \caption{The average reinforced voting weights for each dataset across three random seeds. 
    The values are rounded to four decimal places.}
    \label{tab:weights}
\end{table}

\section{Error Analysis}
\label{app:error_analysis}
\begin{figure*}[t!]\small
    \begin{center}
    \begin{tabular}{@{}c@{}c@{}c@{}c@{}c@{}}
    \includegraphics[width=0.20\textwidth]{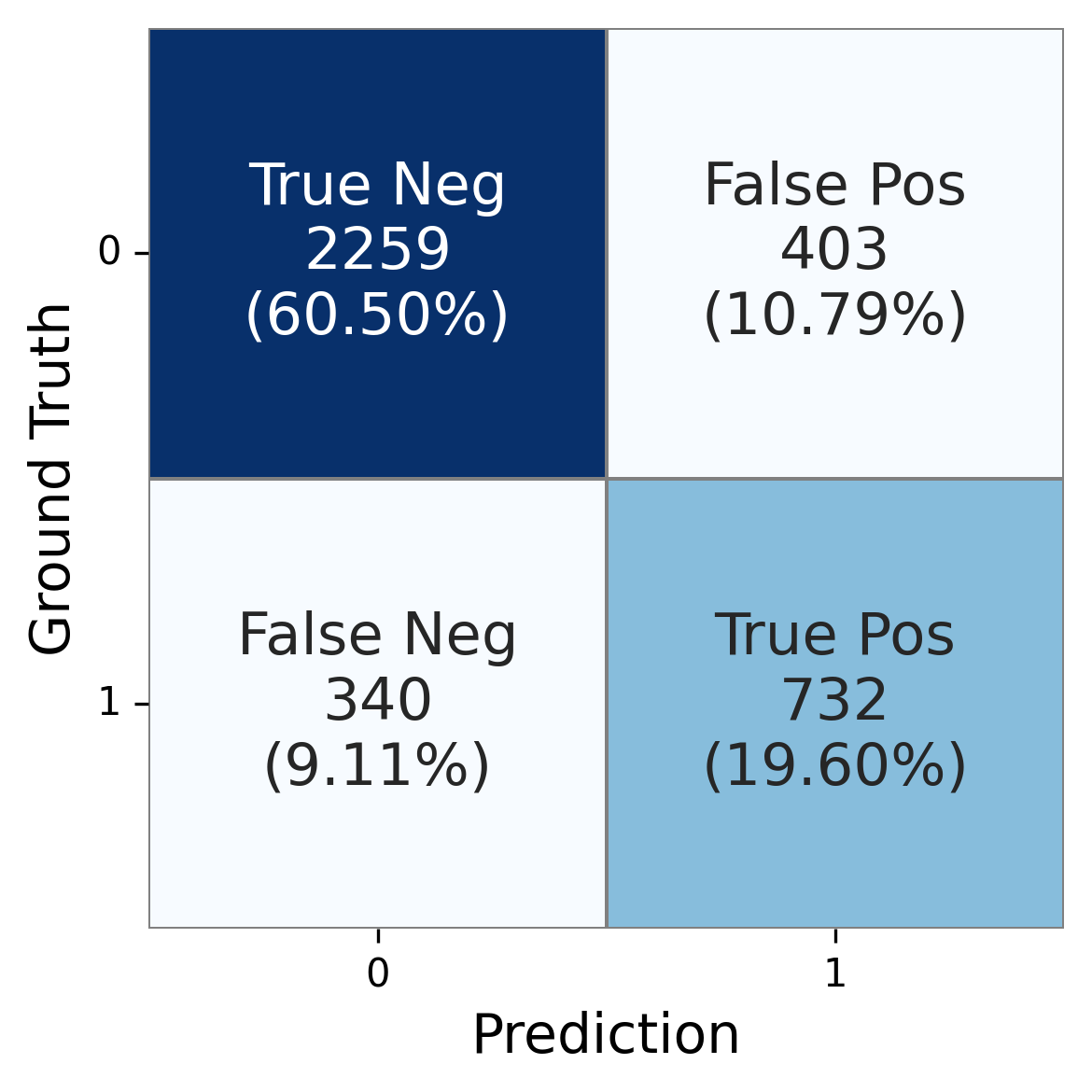} &  
    \includegraphics[width=0.20\textwidth]{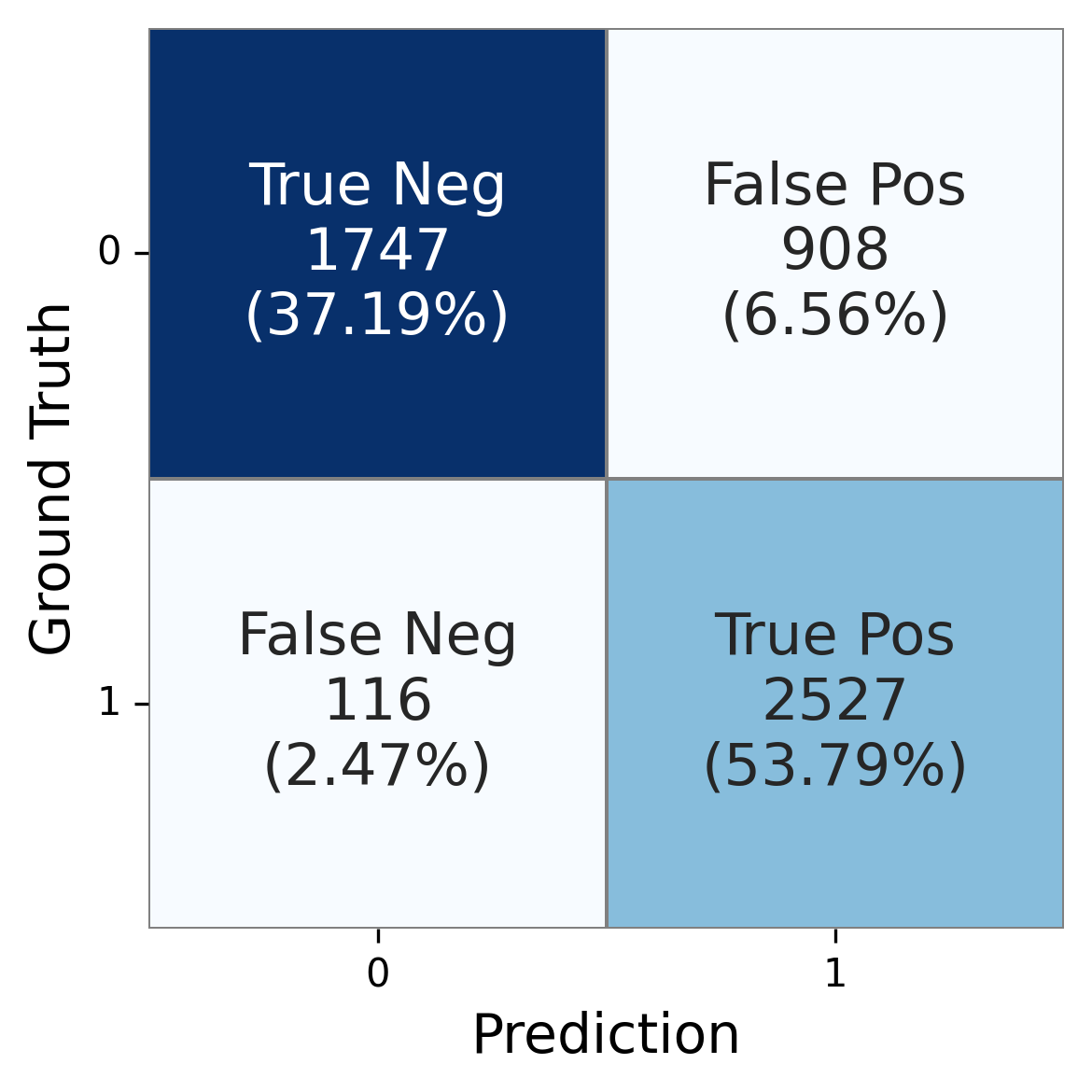} &  
    \includegraphics[width=0.20\textwidth]{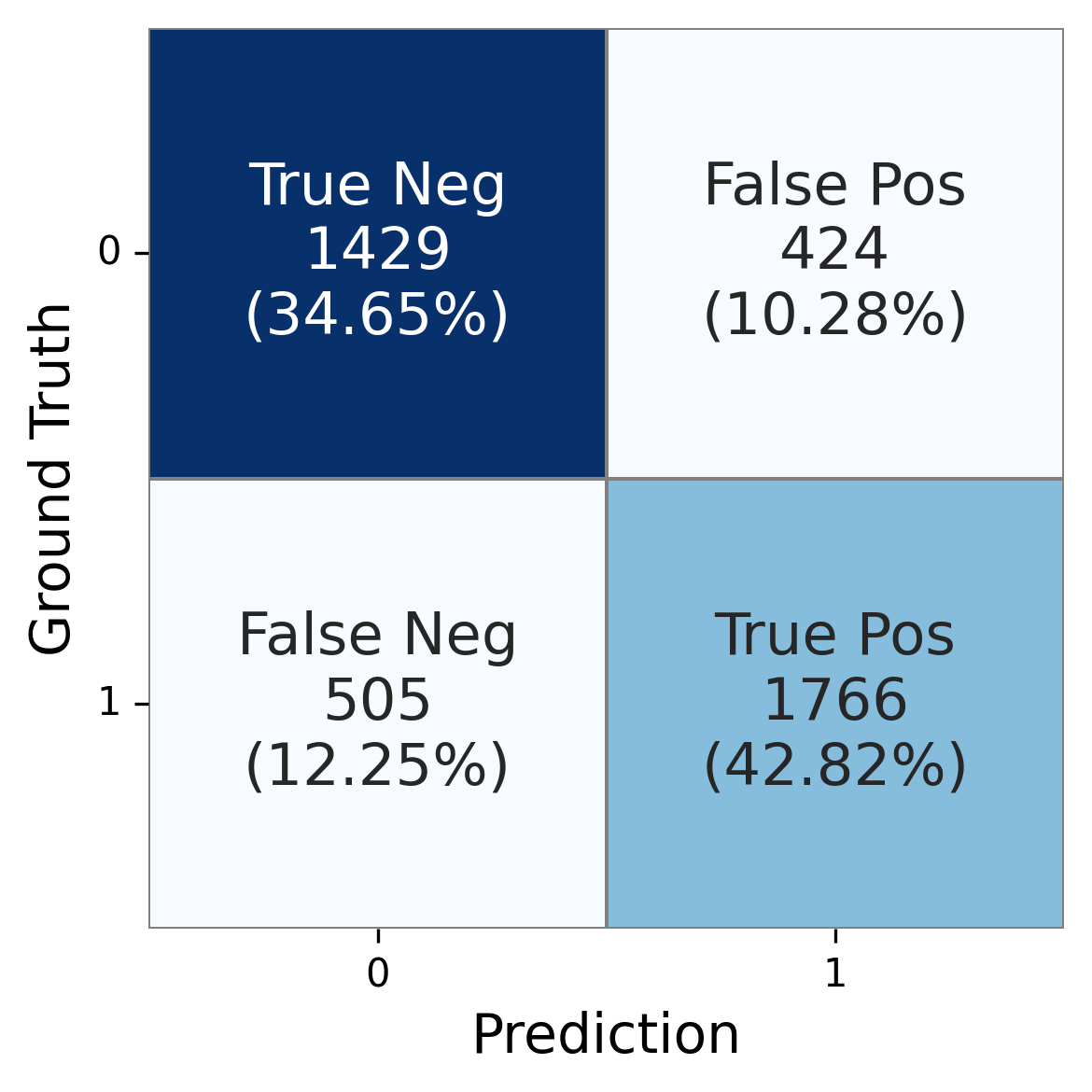} &  
    \includegraphics[width=0.20\textwidth]{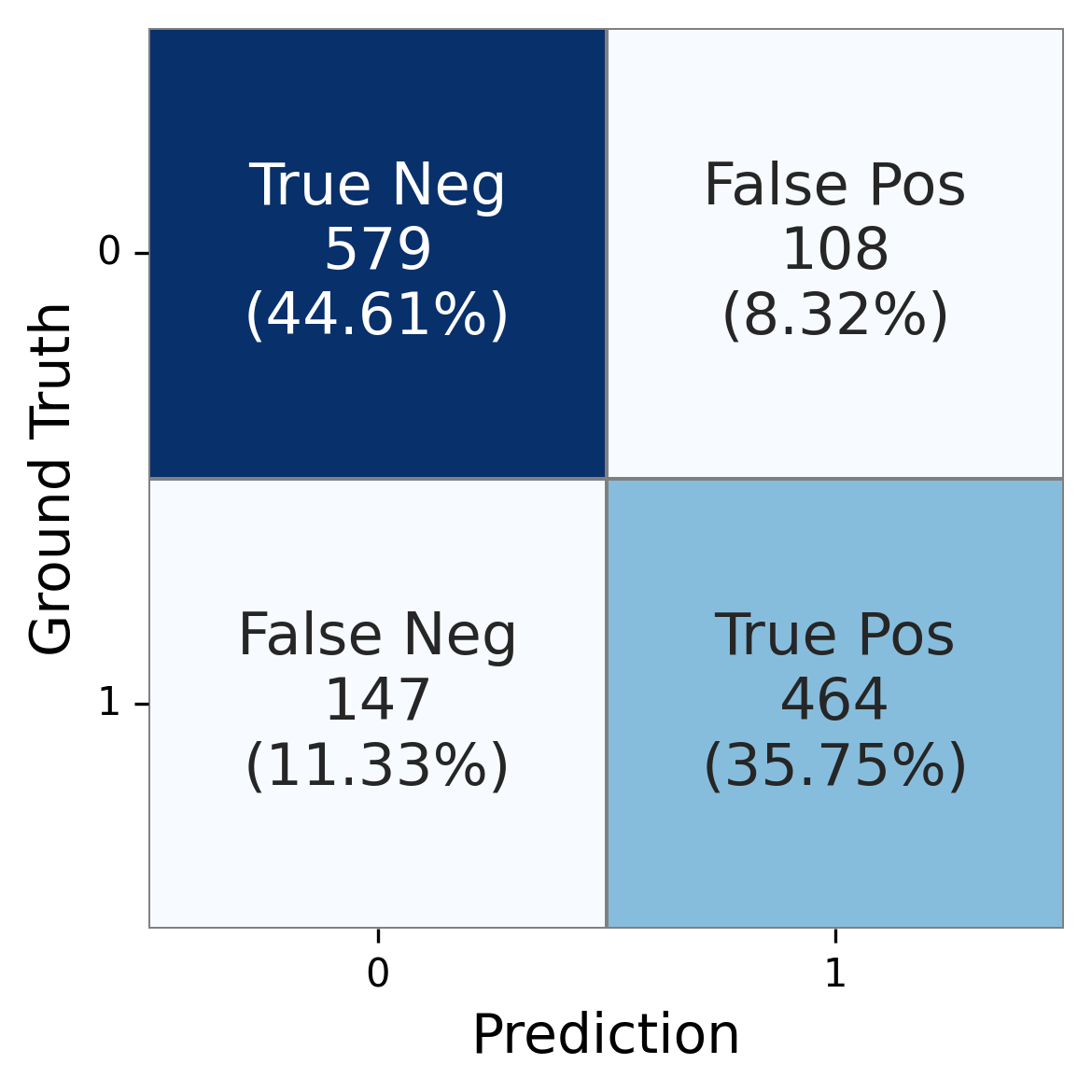} &  
    \includegraphics[width=0.20\textwidth]{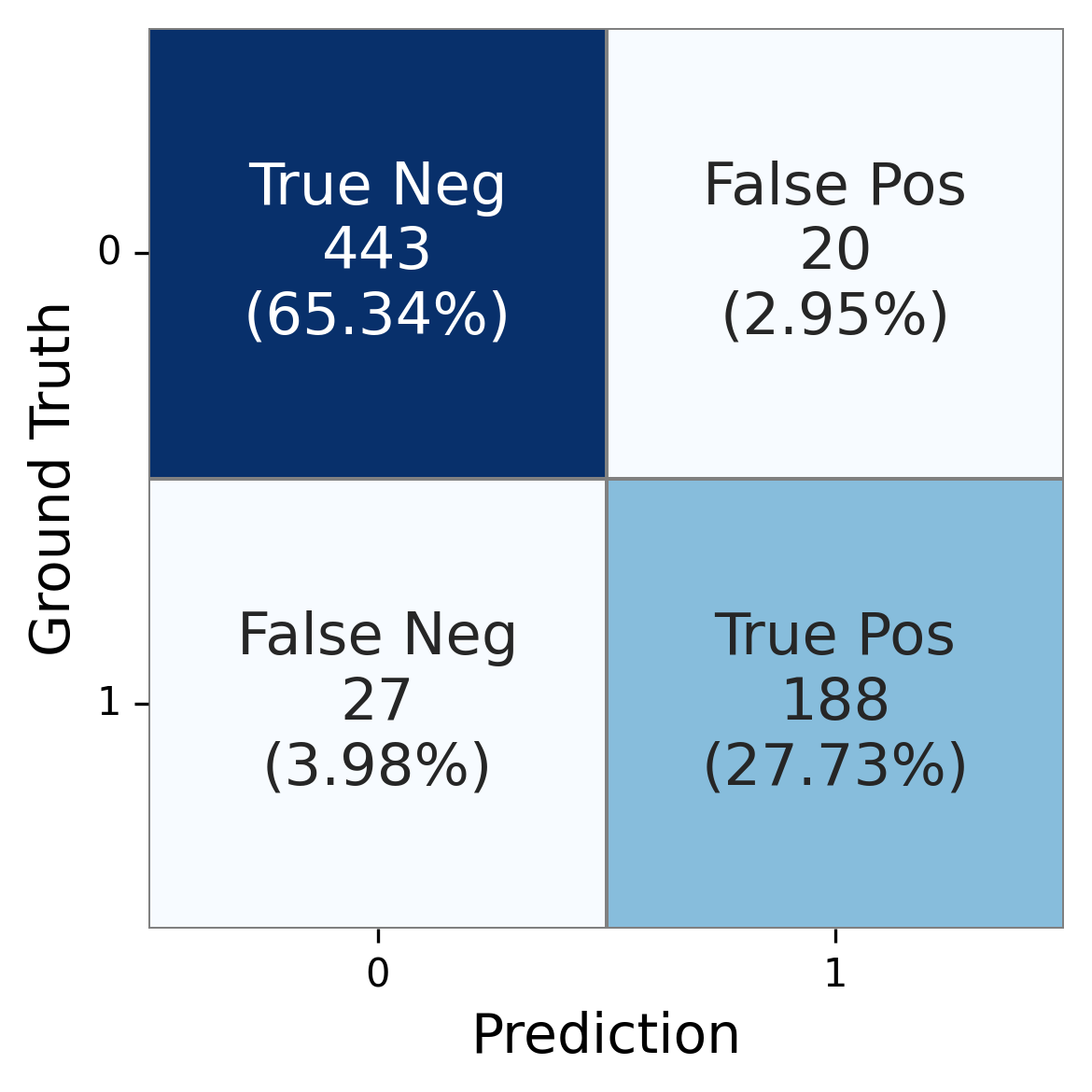} \\ 
    (a) IHC (SharedCon)  & (b) SBIC (SharedCon) & (c) DYNA (SharedCon) & (d) Hateval (SharedCon) & (e) Toxigen (SharedCon)\\
    \includegraphics[width=0.20\textwidth]{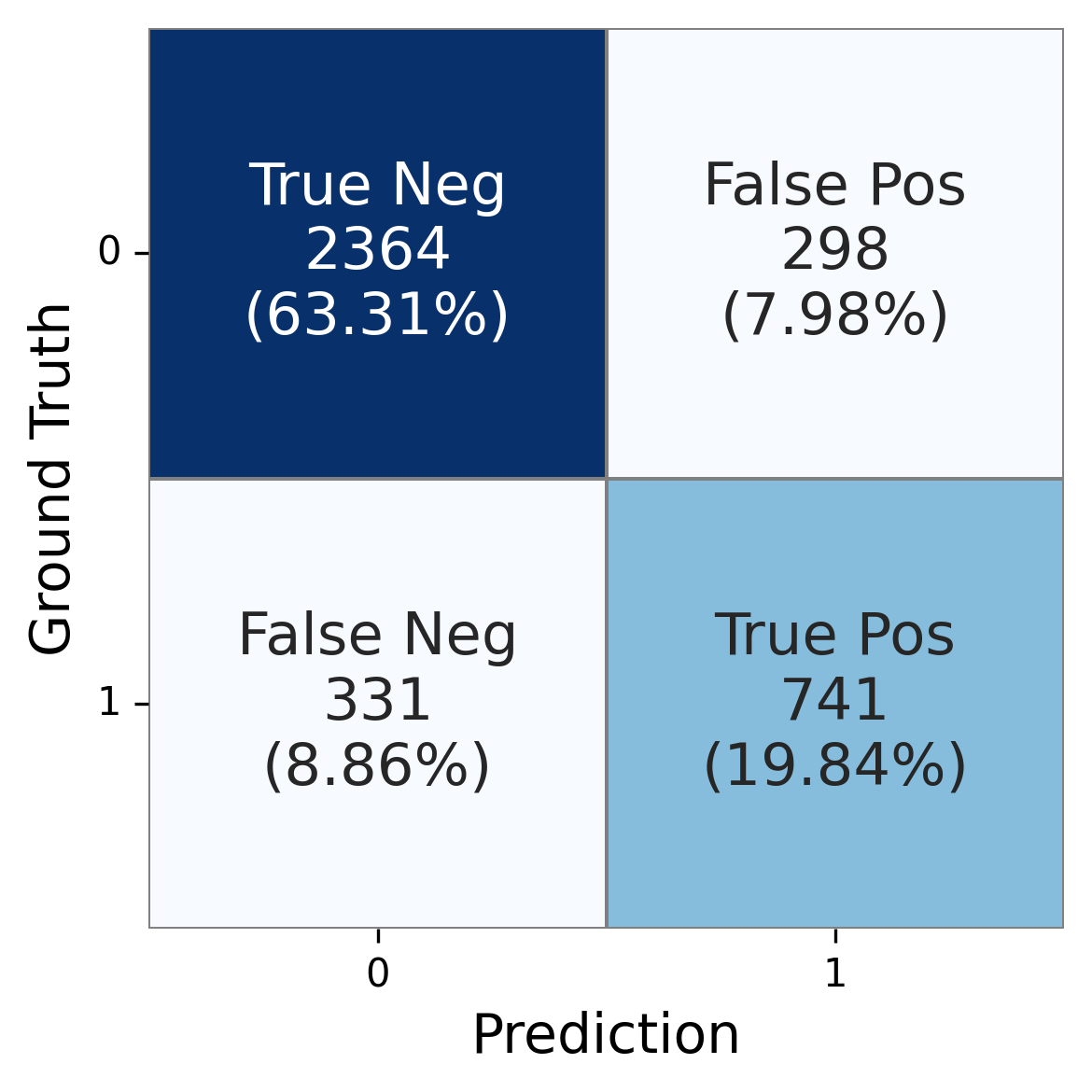} &  
    \includegraphics[width=0.20\textwidth]{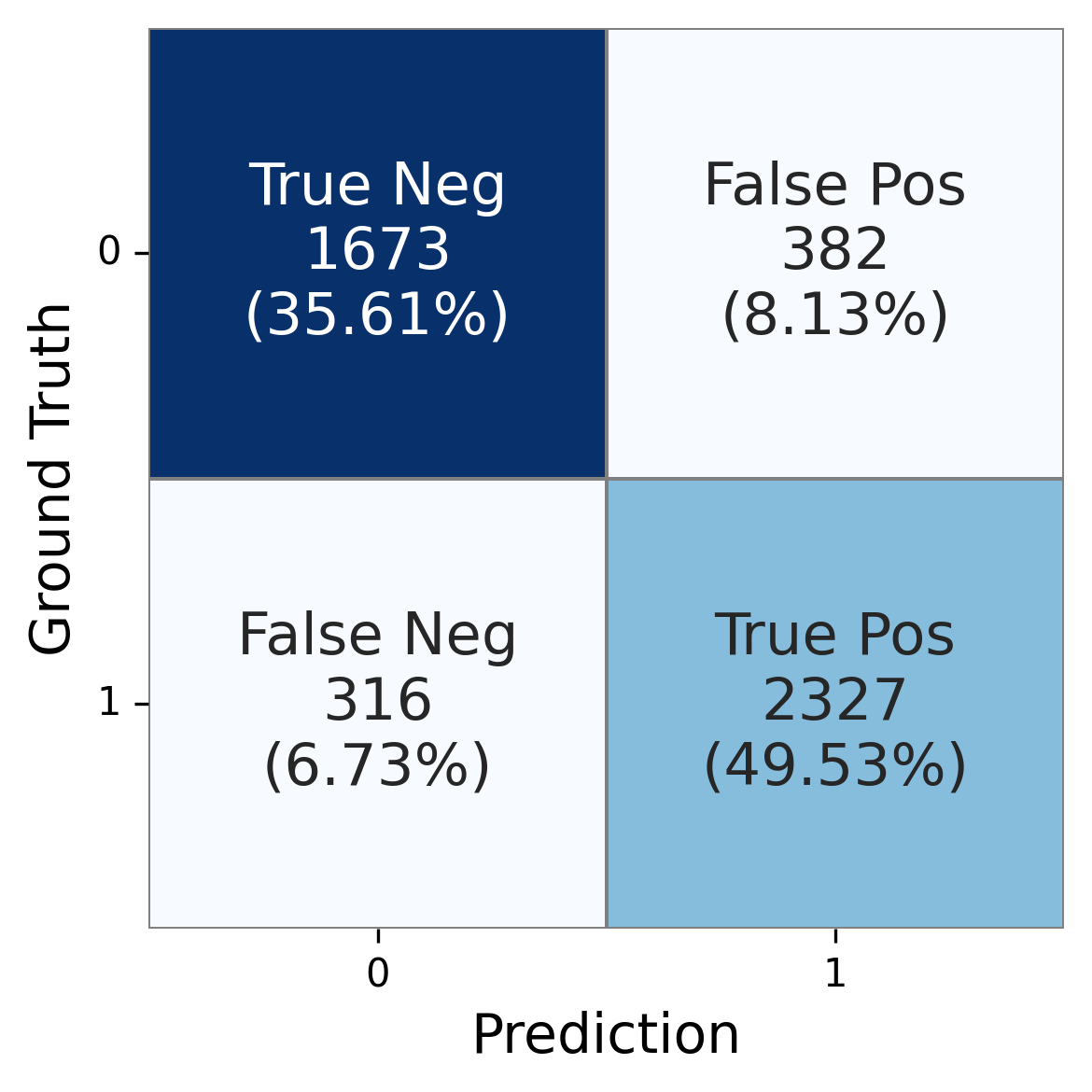} &  
    \includegraphics[width=0.20\textwidth]{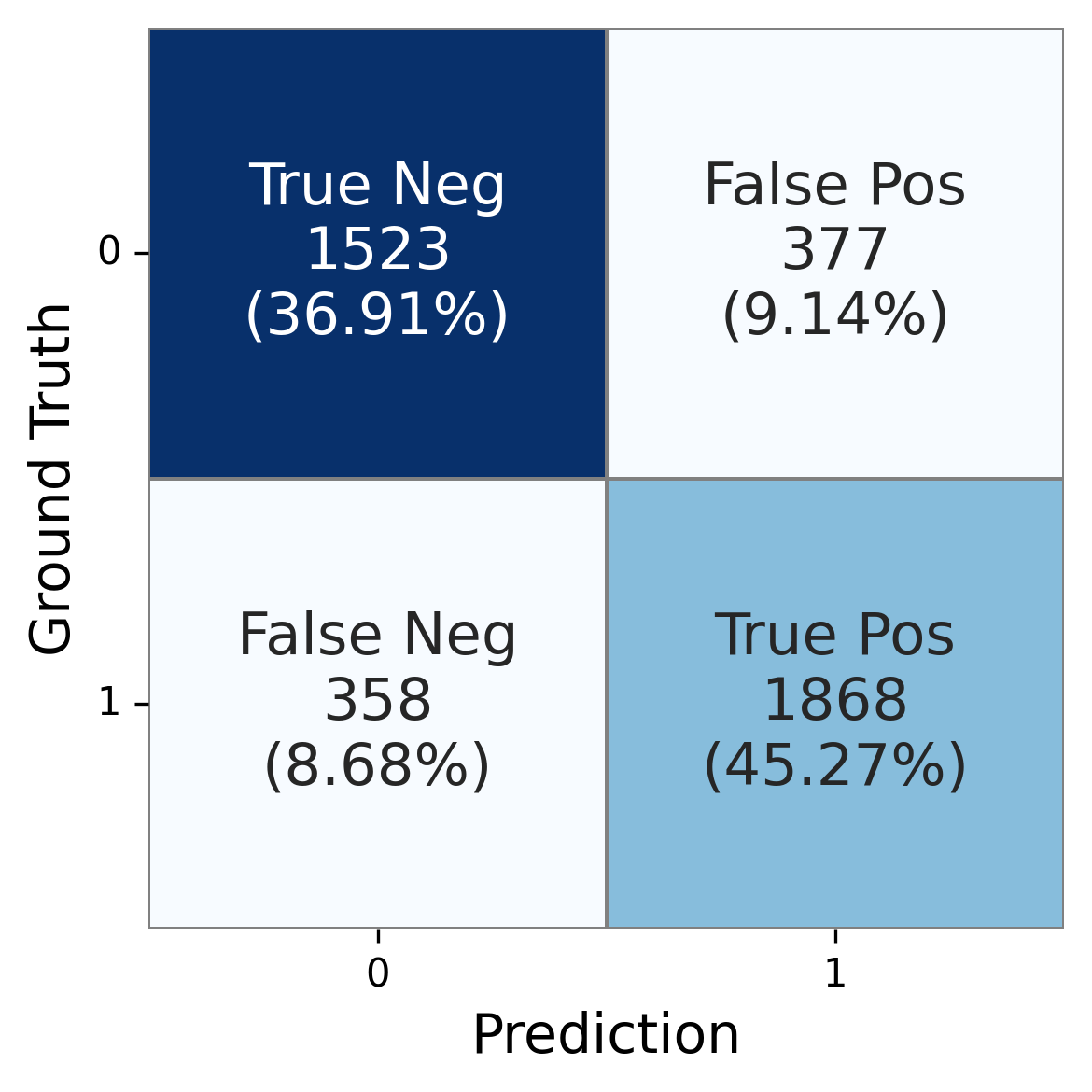} &  
    \includegraphics[width=0.20\textwidth]{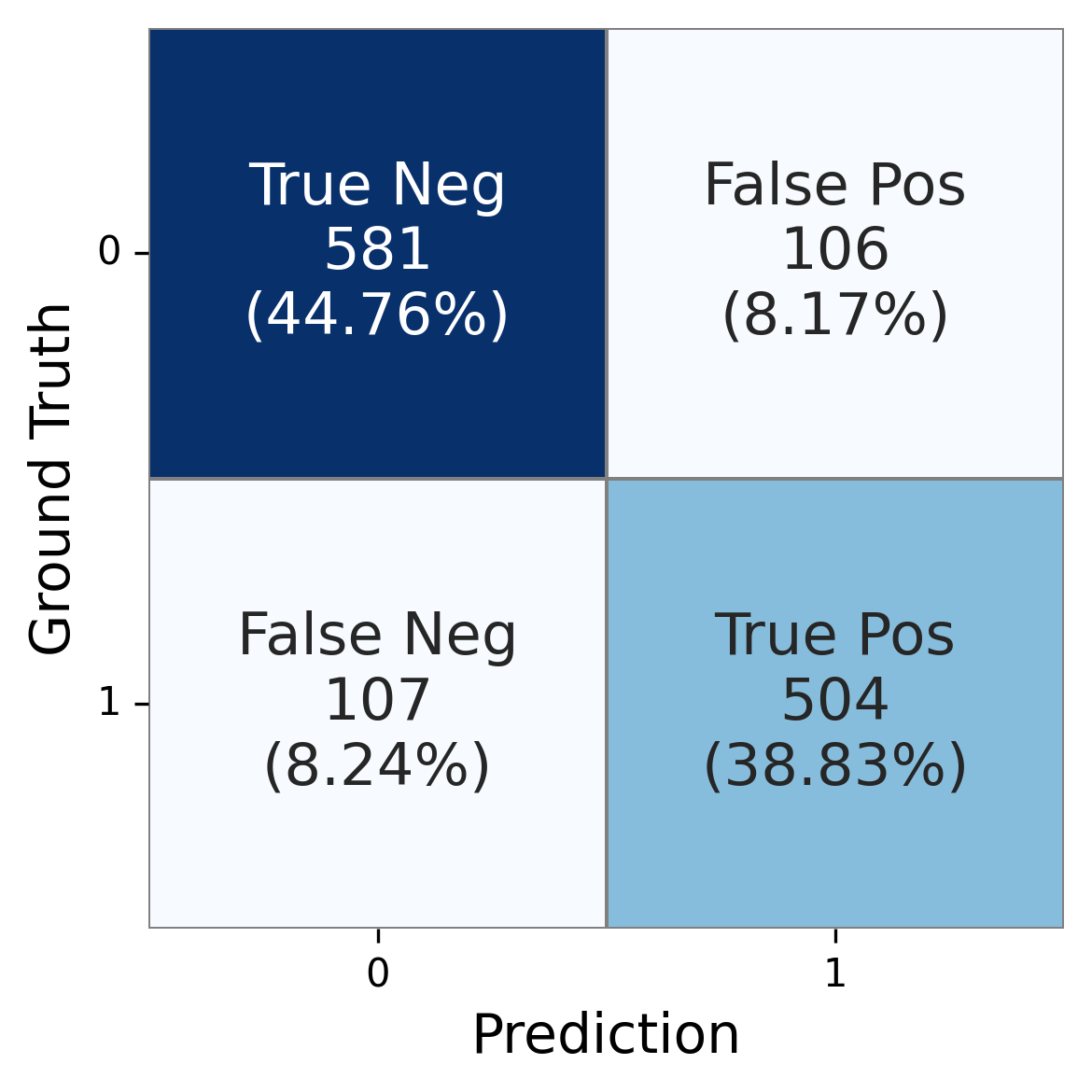} &  
    \includegraphics[width=0.20\textwidth]{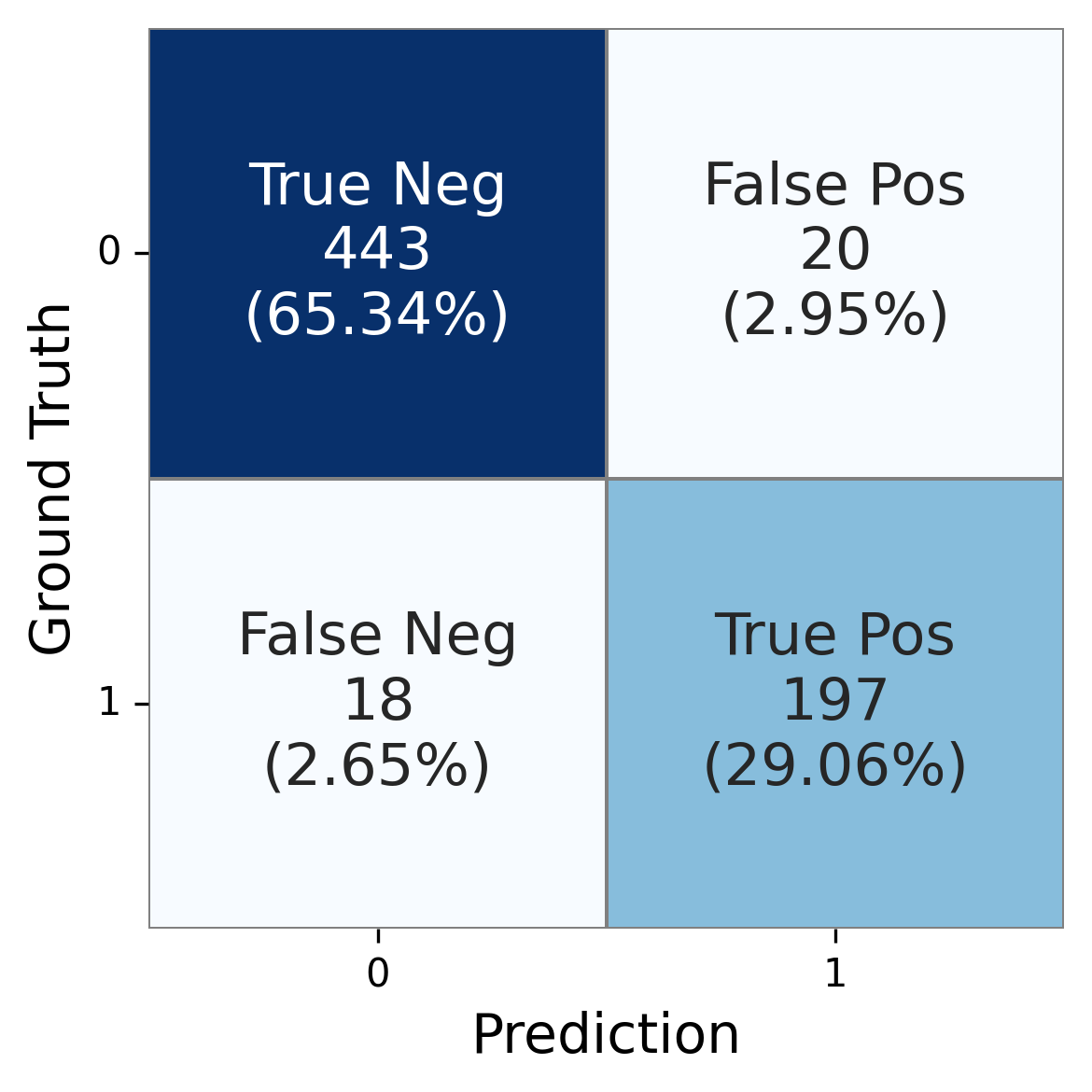} \\ 
    (e) IHC (RV-HATE)  & (f) SBIC (RV-HATE) & (g) DYNA (RV-HATE) & (h) Hateval (RV-HATE) & (e) Toxigen (RV-HATE)\\
    \end{tabular}
    \end{center}
    \caption{Confusion matrices of SharedCon (top row) and RV-HATE (bottom row) on the five hate-speech datasets
    (IHC, SBIC, DYNA, Hateval, and Toxigen). Each cell reports both the absolute count and the percentage of examples
    for true negatives, false positives, false negatives and true positives.}
    \label{fig:confusion}
\end{figure*}

% We analyze the false positive or false negative sampels from each dataset.
% In the case of DYNA, the majority of errors belong to $\mathtt{type2}$ (33.20\%) and
% $\mathtt{type3}$ (44.76\%), with $\mathtt{type3}$ accounting for the largest proportion.
% This result suggests that annotation inconsistencies play a significant role in DYNA errors.
% In particular, high proportion of mislabeled and ambiguous instances indicates
% that DYNA likely contains a variety of dynamic and context-dependent expressions.
% Hateval shows the greatest degree of annotator noise $\mathtt{type3} $(41.78\%), indicating a large number of mislabeled instances.
% % This implies high
% % Toxigen은 데이터 개수가 원래 적어...
% Manual inspection further confirms that the dataset contains substantial label noise.
% In contrast, since Toxigen is a machine-generated dataset,
% it contains relatively few typographical errors
% or broken sentences; 
% note that our analysis identifies zero instances from the error samples.
% However, $\mathtt{type2}$ errors account for 50\% of the error samples,
% indicating that Toxigen includes many semantically ambiguous instances
% despite its clean text.
We provide error examples in Table~\ref{tab:error_types1} and confusion matrices in Figure~\ref{fig:confusion}.
% Table~\ref{tab:error_types2} complements Table~\ref{tab:error_types1} by providing examples from the two remaining
% datasets (Hateval and Toxigen) not reported in the earlier table.
Note that Toxigen does not contain any instances categorized as $\mathtt{type 1}$ error. 
Figure~\ref{fig:confusion} presents a comparison between the confusion matrices of SharedCon,
which serves as both one of our baselines and the previous SOTA model, and our proposed method RV-HATE.
In Hateval, the majority of misclassified examples by RV-HATE fall under $\mathtt{type 3}$ errors.
Upon manual inspection, we observe several cases with highly similar contexts where one instance is
labeled as hate and another as non-hate, suggesting the presence of annotation noise in the dataset.
For Toxigen, the proportion of broken or masked sentences was notably low, likely due to its
synthetic, machine-generated nature. Nevertheless, a substantial number of examples remained ambiguous;
over 50\% of the misclassified samples are categorized as $\mathtt{type 2}$ errors.

% \begin{table*}[h!]
% \centering
% \begin{tabular}{llcp{11cm}l}
% \toprule
% \textbf{Dataset} & \textbf{Error Type} & \textbf{Ratio} & \textbf{Example} & \textbf{Label} \\
% \midrule
% \multirow{3}{*}{Hateval} & $\mathtt{type 1}$ & 25.35\% & correction - hugh if true. \# makemeasandwich & hate \\
%  & $\mathtt{type 2}$ & 27.70\% & when you join a relationship women leave this part out \# yesallmen & hate \\
%  & $\mathtt{type 3}$ & 41.78\% & u a bitch & hate \\
%  & $\mathtt{type 4}$ & 5.16\% & \textemdash \\
% \midrule
% \multirow{3}{*}{Toxigen} & $\mathtt{type 1}$ & 0\% & \textemdash & hate\\
%  & $\mathtt{type 2}$ & 50.00\% & go back to china. & hate\\
%  & $\mathtt{type 3}$ & 21.05\% & ``it is not ok for me to be homophobic, but it is ok for me to be homophobic'' & hate\\
%  & $\mathtt{type 4}$ & 28.95\% & \textemdash \\
% \bottomrule
% \end{tabular}
% \caption{Error types and examples across datasets.}
% \label{tab:error_types2}
% \end{table*}

% Hateval type3이 제일 많고, 그러므로 mislabel이 많이 포함된 데이터인 것을 알 수 있다.
% 또한 Hateval 데이터 자체는 실제 확인결과 label noise가 굉장히 많이 포함되어있는걸 확인하였다. 
% Toxigen은 machine-generated dataset이기 때문에 오타와 masking이 깨진 것들의 비율이 현저히 낮았다.
% 그럼에도 불구하고, ambiguous한 데이터는 많이 존재했는데, error sample중 무려 50\%나 type2였다.
% 따라서 Toxigen은 오타는 적지만, ambiguous한 데이터가 많이 존재하는 데이터셋임을 알 수 있다.

\section{Embeddings}
\label{app:embeddings}
\begin{figure*}[t!]\small
    \begin{center}
    \begin{tabular}{@{}c@{}c@{}c@{}c@{}c@{}c@{}}
    \includegraphics[width=0.2\textwidth]{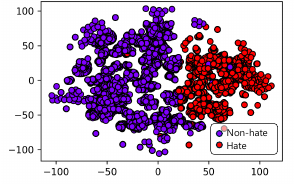} &  
    \includegraphics[width=0.2\textwidth]{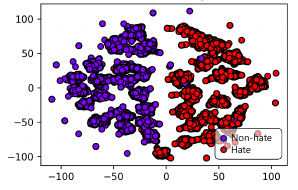} &  
    \includegraphics[width=0.2\textwidth]{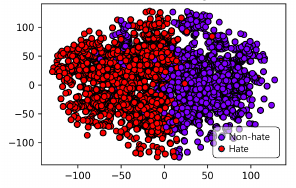} &  
    \includegraphics[width=0.2\textwidth]{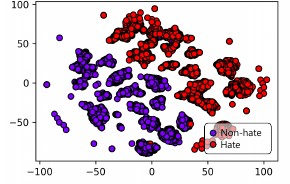} &  
    \includegraphics[width=0.2\textwidth]{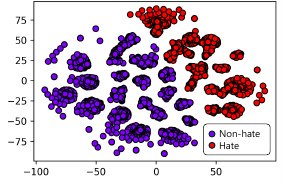}\\  
    (a) IHC ($\mathtt{M_0}$)& (b) SBIC ($\mathtt{M_0}$) & (c) DYNA ($\mathtt{M_0}$) & (d) Hateval ($\mathtt{M_0}$) & (e) Toxigen ($\mathtt{M_0}$)\\
    \includegraphics[width=0.2\textwidth]{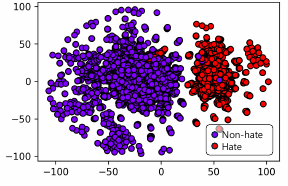} &  
    \includegraphics[width=0.2\textwidth]{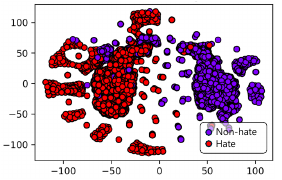} &  
    \includegraphics[width=0.2\textwidth]{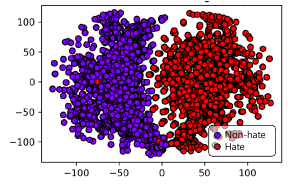} &  
    \includegraphics[width=0.2\textwidth]{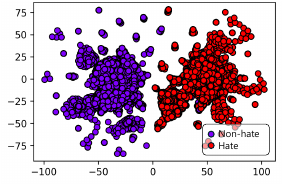} & 
    \includegraphics[width=0.2\textwidth]{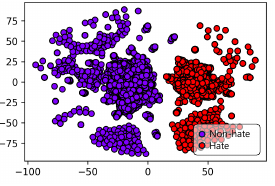} \\ 
    (f) IHC ($\mathtt{M_3}$)& (g) SBIC ($\mathtt{M_3}$) & (h) DYNA ($\mathtt{M_3}$) & (i) Hateval ($\mathtt{M_3}$) & (j) Toxigen ($\mathtt{M_3}$)\\
    \end{tabular}
    \end{center}
    \caption{t-SNE visualization of sentence embeddings from the SBIC, Hateval and Toxigen datasets.
    The top row shows embeddins produced by $\mathtt{M_0}$,  while the bottim row shows those from
    $\mathtt{M_3}$, illustrating the effect of $\mathtt{M_3}$.}
    \label{fig:app_embeddings}
\end{figure*}

Figure~\ref{fig:app_embeddings} visualizes the impact of the hard negative sampling module ($\mathtt{M_3}$)
on the embedding space via t-SNE projections for the SBIC, Hateval and Toxigen datasets.
In the absence of $\mathtt{M_3}$ (top row), embeddings form local clusters but fail to exhibit
a clear separation between `non-hate' class and `hate' class instances. Once $\mathtt{M_3}$
is applied (bottom row), these clusters persist and the two classes become distinctly partitioned,
demonstrating that hard negative sampling sharpens the decision boundary in the representation space.

% \begin{figure}[t!]\small
%     \begin{center}
%     \begin{tabular}{@{}c@{}c@{}c@{}c@{}}
%     \includegraphics[width=0.32\columnwidth]{figures/SBIC_embedding_base.pdf} &  
%     \includegraphics[width=0.32\columnwidth]{figures/Hateval_embedding_base.pdf} &  
%     \includegraphics[width=0.32\columnwidth]{figures/Toxigen_embedding_base.pdf}\\  
%     (a) SBIC ($\mathtt{M_0}$)  & (b) Hateval ($\mathtt{M_0}$) & (c) Toxigen ($\mathtt{M_0}$)\\
%     \includegraphics[width=0.32\columnwidth]{figures/SBIC_embedding_m3.pdf} &  
%     \includegraphics[width=0.32\columnwidth]{figures/Hateval_embedding_m3.pdf} & 
%     \includegraphics[width=0.32\columnwidth]{figures/Toxigen_embedding_m3.pdf} \\ 
%     (d) SBIC ($\mathtt{M_3}$)  & (e) Hateval ($\mathtt{M_3}$) & (f) Toxigen ($\mathtt{M_3}$) \\
%     \end{tabular}
%     \end{center}
%     \caption{t-SNE visualization of sentence embeddings from the SBIC, Hateval and Toxigen datasets.
%     The top row shows embeddins produced by $\mathtt{M_0}$,  while the bottim row shows those from
%     $\mathtt{M_3}$, illustrating the effect of $\mathtt{M_3}$.}
%     \label{fig:app_embeddings}
% \end{figure}

\section{IQR}
\label{app:iqr}

The IQR is defined as the difference between the third quartile ($Q_3$) and 
the first quartile ($Q_1$).
The upper bound threshold is calculated by adding 1.5 times the IQR to the $Q_3$.
\begin{equation}
IQR = Q_3 - Q_1.
\end{equation}
\begin{equation}
upper\_bound=Q_3 + 1.5 \times IQR.
\end{equation}
The resulting set of data points after removing outliers $X$ includes only
The resulting set of data points after removing outliers includes only
points whose distances from the cluster center are less than the upper bound.
\begin{equation}
X = \{x|x<upper\_bound\}.
\end{equation}

\section{Policy Network Architecture}
\label{app:ppo}
We describe the architecture of the policy network used for reinforcement learning in RV-HATE.
The policy network takes a four-dimensional state as input and outputs a weight vector $w=[w_0, w_1,w_2,w_3]$, where each weight corresponds to a module in the voting process. The network consists of a shred feature extractor followed by separate actor and critic heads.

The shared feature extractor is implemented as a row-layer feedforward network with hidden dimension 64. 
Give an input state $s\in\mathbb{R}^4$, the hidden representation is computed through tow linear layers with nonlinear activation.
This component contains 4,480 parameters.

The actor head maps the shared representation to a four-dimensional action space and contains 260 parameters.
The critic head outputs a scalar value estimate and contains 65 parameters.
In total, the policy network contains 4,805 parameters,
which is negligible compared to the BERT-base encoders used in each module.

\section{Ablation}
\label{app:ablation-analysis}
\begin{figure*}[h!]\small
  \centering
  % (a)
  \begin{subfigure}[b]{0.49\linewidth}
    \includegraphics[width=\linewidth]{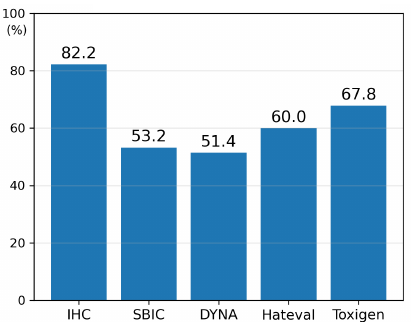}
    \caption{Implicit hate speech ratio}
    \label{fig:ratio:implicit}
  \end{subfigure}
  \hfill
  % (b)
  \begin{subfigure}[b]{0.49\linewidth}
    \includegraphics[width=\linewidth]{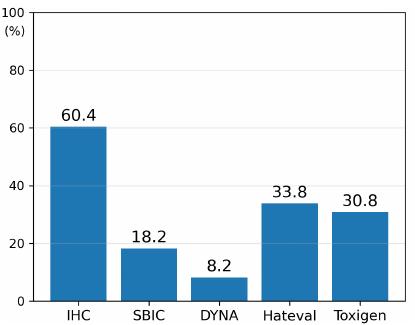}
    \caption{Broken sentence ratio}
    \label{fig:ratio:broken}
  \end{subfigure}
  \caption{The ratio of implicit hate speech and broken sentence for each dataset we used.}
  \label{fig:ablation_ratio}
\end{figure*}

Figure~\ref{fig:ablation_ratio} presents the results of the ablation
study. Subfigure (a) shows the ratio of implicit hate speech,
which demonstrates the effectiveness of $\mathtt{M_1}$,
while subfigure (b) illustrates the distribution of broken sentences,
highlighting the impact of $\mathtt{M_2}$.
Except for Toxigen, all experiments are conducted on 500 randomly
sampled instances from each dataset, initially annotated using gpt-4.1 and subsequently verified by two experts. 
Figure~\ref{fig:prompt_broken_sentence} and ~\ref{fig:prompt_implicit} present the prompts used to identify broken sentences and instances
of implicit hate speech, respectively.

\begin{figure*}[h!]
\begin{tcolorbox}[colback=gray!10,colframe=black,width=\textwidth]
You are a broken sentence detect GPT.
When given a sentence, follow the instructions below:

1. The provided sentences are examples of hate speech. \\
2. A broken sentence is an incomplete or fragmented sentence that lacks essential grammatical components.\\
3. If a sentence contains a typo, it is also regarded as a brokensentence.\\
4. Sentences that include hashtags or a large number of special characters are also considered broken sentences.\\
5. Determine whether the given sentence is broken sentence or not.\\
6. Ouput 1 if borken sentence, and 0 if it's not.\\

\textbf{Example 1}\\
Given sentence: An immigrant wouldn't ask that question if his familiy had been murdered by savages.\\
Your response: 0\\

\textbf{Example 2}\\
Given sentence: Awwwwww they no Kaep aint rape no bitch!!!!\\
Your response: 1\\

\textbf{Example 3}\\
Given sentence: @StephyRae7 hii fag. :)\\
Your response: 1\\

Remember: You have to print only 0 or 1. As shown in the example, no additional explanation is necessary in your response.

\medskip
\hrule
\medskip
Post: \texttt{[Sentence]}
\end{tcolorbox}
\caption{Prompt for broken sentence}
\label{fig:prompt_broken_sentence}
\end{figure*}

\begin{figure*}[h!]
\begin{tcolorbox}[colback=gray!10,colframe=black,width=\textwidth]
You are a implicit hate detect GPT.
When given a sentence, follow the instructions below:

1. The provided sentences are examples of hate speech. \\
2. Hate speech consists of two types.\\
3. Explicit hate is hate data that directly express hatred.\\
4. Implicit hate is hate data that convey hatred indirectly through context.\\
5. Determine whether the given sentence is explicit hate or implicit hate.\\
6. Ouput 0 if it's explicit, and 1 if it's implicit.\\

[notice]\\
- You have to print only 0 or 1
- As shown in the example, no additional explanation is necessary in your response. \\

\textbf{Example}\\
Given sentence: immigrants wouldn't ask that question if his family was murdered by savages\\
Your response: 1\\

\medskip
\hrule
\medskip
Post: \texttt{[Sentence]}
\end{tcolorbox}
\caption{Prompt for implicit hate speech}
\label{fig:prompt_implicit}
\end{figure*}

\end{document}